%% file: main.tex
\documentclass[10pt,twocolumn,letterpaper]{article}
\usepackage{cvpr}
\usepackage{times}
\usepackage{epsfig}
\usepackage{graphicx}
\usepackage{amsmath}
\usepackage{subcaption}
\usepackage{amssymb}
\usepackage{math_style}
\usepackage[normalem]{ulem}
\usepackage[usenames,dvipsnames,svgnames,table]{xcolor}
\usepackage{multirow}
\usepackage{tikz} 
\usepackage{balance}
\usepackage{pgfplots}
\usetikzlibrary{shapes.geometric}
\usepackage{algorithm}
\usepackage{algpseudocode}
\usepackage{siunitx}

\newcolumntype{C}[1]{>{\centering\arraybackslash}p{#1}}

\newbox\mybox
\def\centerfigure#1{%
    \setbox\mybox\hbox{#1}%
    \raisebox{-0.5\dimexpr\ht\mybox+\dp\mybox}{\copy\mybox}%
}

\graphicspath{images}

\usepackage[pagebackref=true,breaklinks=true,letterpaper=true,colorlinks,bookmarks=false]{hyperref}

\newcommand{\ARXIV}[2]{#1} % for ARXIV
\cvprfinalcopy % *** Uncomment this line for the final submission

\def\cvprPaperID{1915} % *** Enter the CVPR Paper ID here

\date{} %<-remove for CVPR
\begin{document}
%%%%%%%%% TITLE
\title{Large-scale Point Cloud Semantic Segmentation with Superpoint Graphs}
\author{Loic Landrieu$^{1\star}$, Martin Simonovsky$^{2\star}$\\
$^1$~Universit\'e Paris-Est, LASTIG MATIS IGN, ENSG\\
$^2$~Universit\'e Paris-Est, Ecole des Ponts ParisTech\\
{\tt\small loic.landrieu@ign.fr, martin.simonovsky@enpc.fr}}
% For a paper whose authors are all at the same institution,
% omit the following lines up until the closing ``}''.
% Additional authors and addresses can be added with ``\and'',
% just like the second author.
% To save space, use either the email address or home page, not both
%}
\maketitle
%\ifcvprfinal\thispagestyle{empty}\fi
\let\thefootnote\relax\footnote{$^\star$ Both authors contributed equally to this work.}
%%%%%%%%% ABSTRACT
\begin{abstract} 
We propose a novel deep learning-based framework to tackle the challenge of semantic segmentation of large-scale point clouds of millions of points.
We argue that the organization of 3D point clouds can be efficiently captured by a structure called superpoint graph (SPG), derived from a partition of the scanned scene into geometrically homogeneous elements. SPGs offer a compact yet rich representation of contextual relationships between object parts, which is then exploited by a graph convolutional network. Our framework sets a new state of the art for segmenting outdoor LiDAR scans ($+11.9$ and $+8.8$ mIoU points for both Semantic3D test sets), as well as indoor scans ($+12.4$ mIoU points for the S3DIS dataset).
\end{abstract} 
%%%%%%%%% BODY TEXT

%=================================================
\section{Introduction}
%=================================================

Semantic segmentation of large 3D point clouds presents numerous challenges, the most obvious one being the scale of the data. Another hurdle is the lack of clear structure akin to the regular grid arrangement in images.
These obstacles have likely prevented Convolutional Neural Networks (CNNs) from achieving on irregular data the impressive performances attained for speech processing or images.

Previous attempts at using deep learning for large 3D data were trying to replicate successful CNN architectures used for image segmentation. For example, SnapNet \cite{boulch2017unstructured} converts a 3D point cloud into a set of virtual 2D RGBD snapshots, the semantic segmentation of which can then be projected on the original data. SegCloud \cite{tchapmi2017segcloud} uses 3D convolutions on a regular voxel grid. However, we argue that such methods do not capture the inherent structure of 3D point clouds, which results in limited discrimination performance. Indeed, converting point clouds to 2D format comes with loss of information and requires to perform surface reconstruction, a problem arguably as hard as semantic segmentation. Volumetric representation of point clouds is inefficient and tends to discard small details.

Deep learning architectures specifically designed for 3D point clouds \cite{qi2016pointnet, simonovsky2017dynamic,Riegler2017OctNet,QiYSG17PointNetPP,Engelmann17_3dsemseg} display good results, but are limited by the size of inputs they can handle at once.

\begin{figure*}[t!]
\begin{tabular}{cccc}
	\begin{subfigure}[b]{0.24\textwidth}
		\includegraphics[width=1\textwidth]{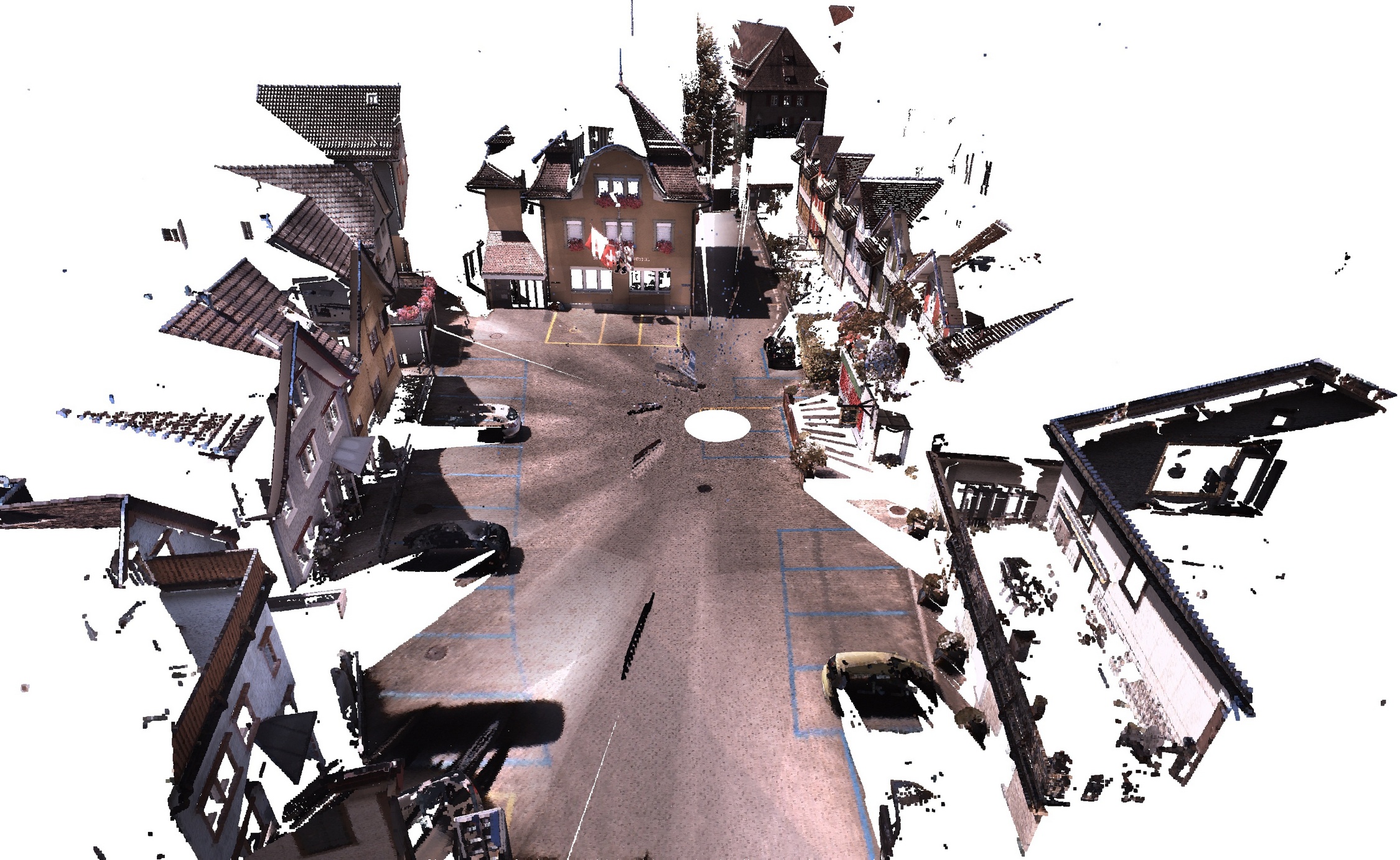}
        \caption{RGB point cloud}
        \label{fig:illu_rgb}
	\end{subfigure}&
    \begin{subfigure}[b]{0.24\textwidth}
		\includegraphics[width=1\textwidth]{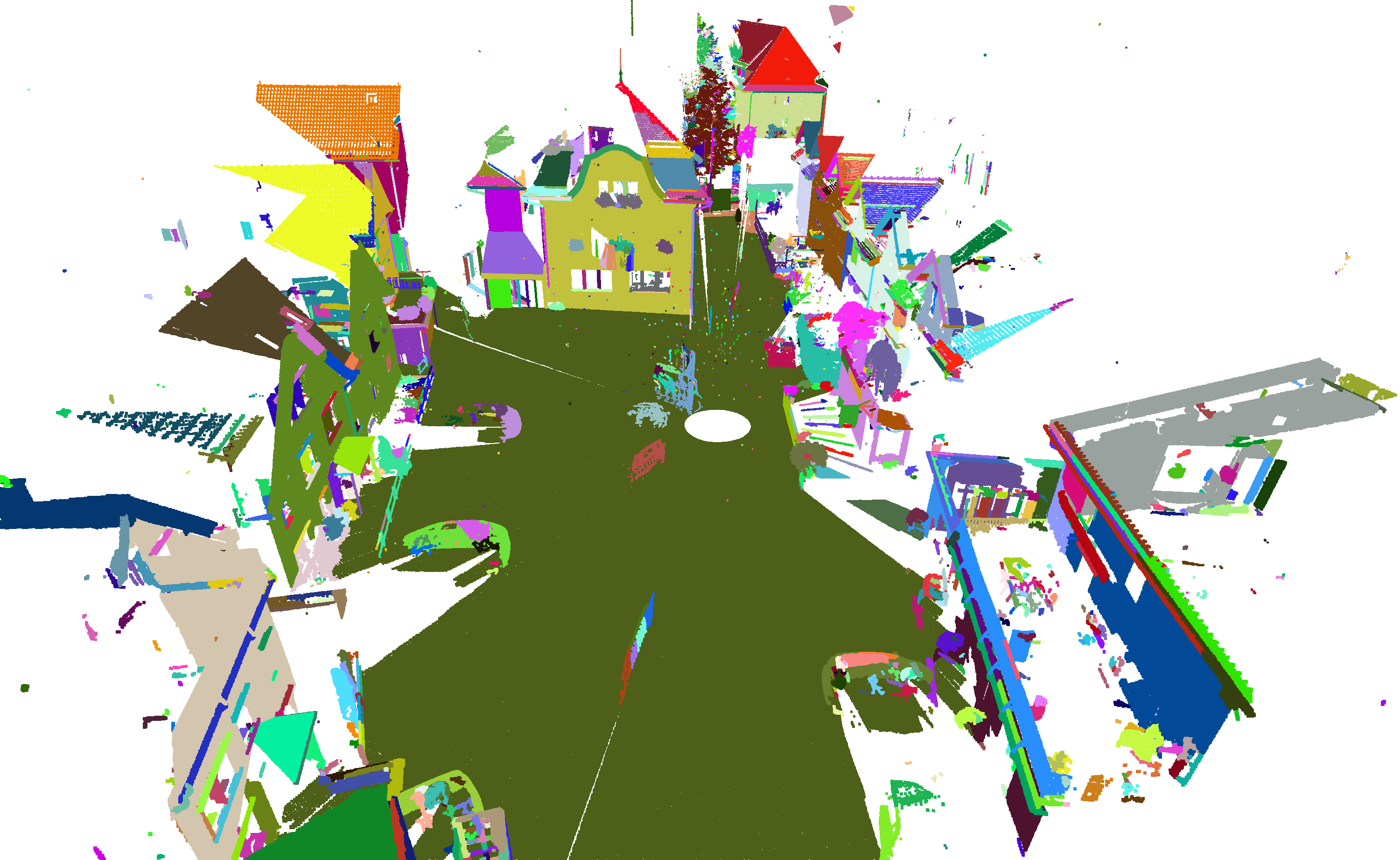}
        \caption{Geometric partition}
        \label{fig:illu_seg}
	\end{subfigure}&
     \begin{subfigure}[b]{0.24\textwidth}   
		\includegraphics[width=1\textwidth]{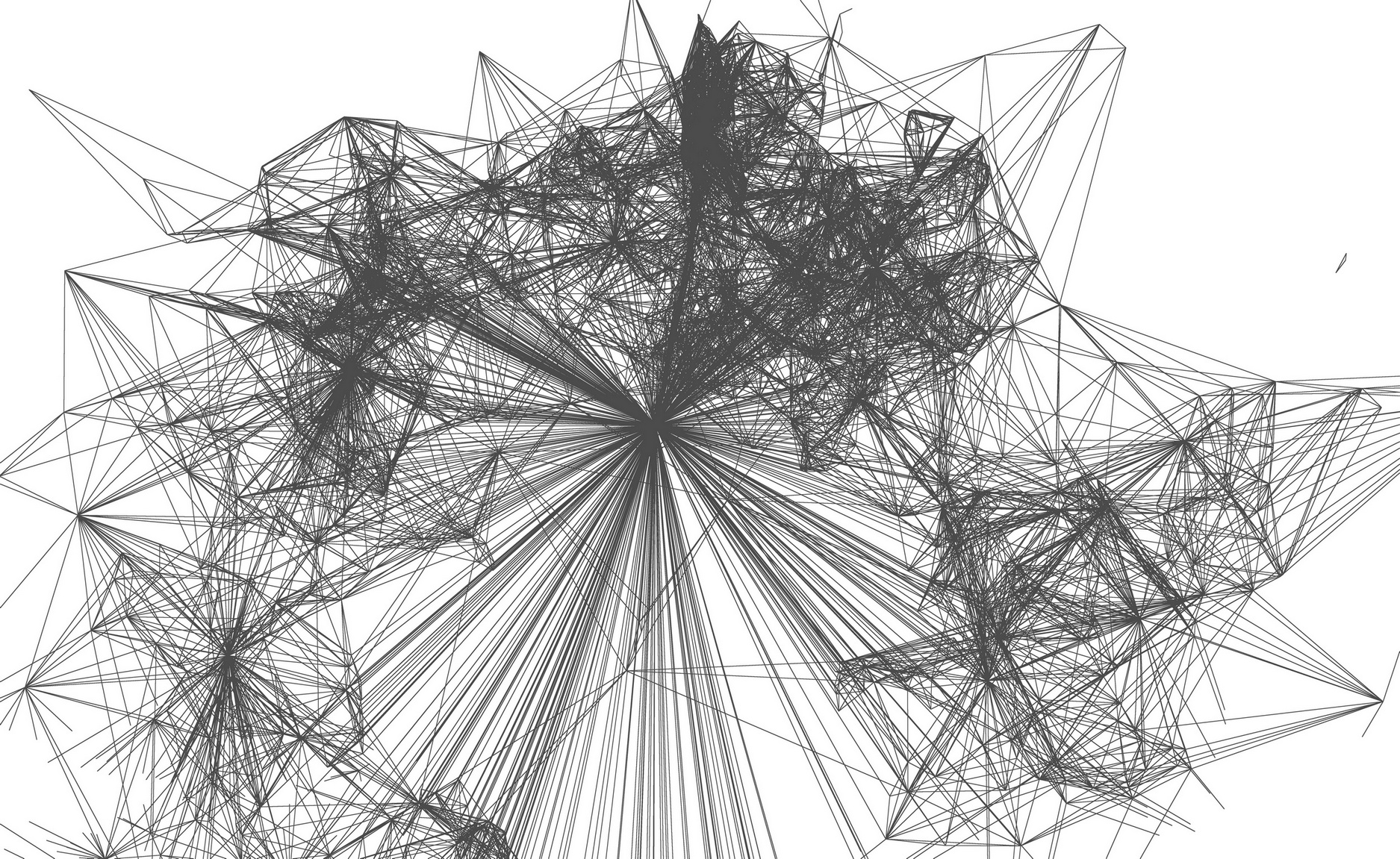}
          \caption{Superpoint graph}
           \label{fig:illu_graph} 
	\end{subfigure}&
    \begin{subfigure}[b]{0.24\textwidth}
		\includegraphics[width=1\textwidth]{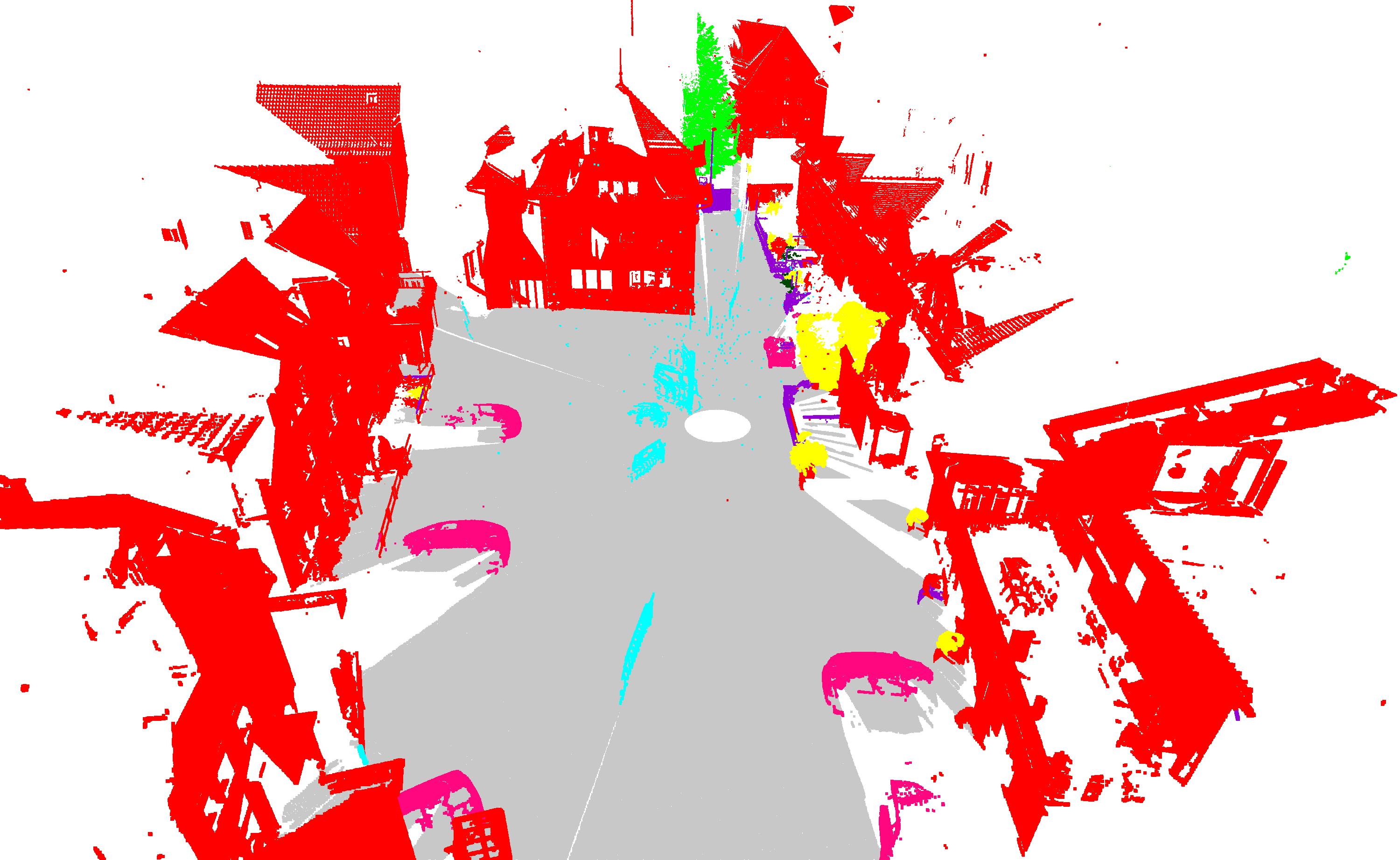}       
        \caption{Semantic segmentation}
         \label{fig:illu_pred}
	\end{subfigure}
\end{tabular}
\caption{Visualization of individual steps in our pipeline. An input point cloud \Subref{fig:illu_rgb} is partitioned into geometrically simple shapes, called superpoints \Subref{fig:illu_seg}. Based on this preprocessing, a superpoints graph (SPG) is constructed by linking nearby superpoints by superedges with rich attributes \Subref{fig:illu_graph}. Finally, superpoints are transformed into compact embeddings, processed with graph convolutions to make use of contextual information, and classified into semantic labels.}
\label{fig:teaser}
\end{figure*}

We propose a representation of large 3D point clouds as a collection of interconnected simple shapes coined superpoints, in spirit similar to superpixel methods for image segmentation \cite{achanta2012slic}. As illustrated in \figref{fig:teaser}, this structure can be captured by an attributed directed graph called the superpoint graph (SPG). Its nodes represent simple shapes while edges describe their adjacency relationship characterized by rich edge features.

The SPG representation has several compelling advantages. First, instead of classifying individual points or voxels, it considers entire object parts as whole, which are easier to identify. Second, it is able to describe in detail the relationship between adjacent objects, which is crucial for contextual classification: cars are generally above roads, ceilings are surrounded by walls, etc. Third, the size of the SPG is defined by the number of simple structures in a scene rather than the total number of points, which is typically several order of magnitude smaller. This allows us to model long-range interaction which would be intractable otherwise without strong assumptions on the nature of the pairwise connections.
%*************************************************
%*************************************************
Our contributions are as follows:
\begin{itemize}
\item We introduce superpoint graphs, a novel point cloud representation with rich edge features encoding the contextual relationship between object parts in 3D point clouds.
\item Based on this representation, we are able to apply deep learning on large-scale point clouds without major sacrifice in fine details. Our architecture consists of PointNets~\cite{qi2016pointnet} for superpoint embedding and graph convolutions for contextual segmentation. For the latter, we introduce a novel, more efficient version of Edge-Conditioned Convolutions~\cite{simonovsky2017dynamic} as well as a new form of input gating in Gated Recurrent Units~\cite{cho-gru14}.
\item We set a new state of the art on two publicly available datasets: Semantic3D \cite{hackel2017semantic3d} and S3DIS \cite{armeni_cvpr16}. In particular, we improve mean per-class intersection over union (mIoU) by $11.9$ points for the Semantic3D reduced test set, by $8.8$ points for the Semantic3D full test set, and by up to $12.4$ points for the S3DIS dataset.
\end{itemize}

%-------------------------------------------------
\section{Related Work}
%-------------------------------------------------
The classic approach to large-scale point cloud segmentation is to classify each point or voxel independently using handcrafted features derived from their local neighborhood \cite{weinmann_contextual_2015}.
The solution is then spatially regularized using graphical models \cite{munoz2009contextual, koppula2011semantic, lu2012simplified, shapovalov2013spatial, kim20133d, anand2013contextually, niemeyer2014contextual,martinovic20153d,wolf2015fast} or structured optimization \cite{LANDRIEU2017102}.
Clustering as  preprocessing \cite{hu2013efficient,guinard_weakly_2017} or postprocessing \cite{weinmann2017hybrid} have been used by several frameworks to improve the accuracy of the classification.

\textbf{Deep Learning on Point Clouds.} Several different approaches going beyond naive volumetric processing of point clouds have been proposed recently, notably set-based~\cite{qi2016pointnet,QiYSG17PointNetPP}, tree-based~\cite{Riegler2017OctNet,KlokovL17KdNet}, and graph-based~\cite{simonovsky2017dynamic}. However, very few methods with deep learning components have been demonstrated to be able to segment large-scale point clouds. PointNet~\cite{qi2016pointnet} can segment large clouds with a sliding window approach, therefore constraining contextual information within a small area only. Engelmann \etal~\cite{Engelmann17_3dsemseg} improves on this by increasing the context scope with multi-scale windows or by considering directly neighboring window positions on a voxel grid. SEGCloud~\cite{tchapmi2017segcloud} handles large clouds by voxelizing followed by interpolation back to the original resolution and post-processing with a conditional random field (CRF). None of these approaches is able to consider fine details and long-range contextual information simultaneously. In contrast, our pipeline partitions point clouds in an adaptive way according to their geometric complexity and allows deep learning architecture to use both fine detail and interactions over long distance.

\textbf{Graph Convolutions.} A key step of our approach is using graph convolutions to spread contextual information. Formulations that are able to deal with graphs of variable sizes can be seen as a form of message passing over graph edges~\cite{GilmerSRVD17}. Of particular interest are models supporting continuous edge attributes~\cite{simonovsky2017dynamic,MontiBMRSB17}, which we use to represent interactions. In image segmentation, convolutions on graphs built over superpixels have been used for post-processing: Liang \etal~\cite{LiangSFLY16,LiangLSFYX17} traverses such graphs in a sequential node order based on unary confidences to improve the final labels. We update graph nodes in parallel and exploit edge attributes for informative context modeling. Xu \etal~\cite{Xu17GraphGen} convolves information over graphs of object detections to infer their contextual relationships. Our work infers relationships implicitly to improve segmentation results. Qi \etal~\cite{QiLJFU17} also relies on graph convolutions on 3D point clouds. However, we process large point clouds instead of small RGBD images with nodes embedded in 3D instead of 2D in a novel, rich-attributed graph. Finally, we note that graph convolutions also bear functional similarity to deep learning formulations of CRFs~\cite{Zheng15crf}, which we discuss more in \secref{subsec:contexseg}.
%=================================================
\section{Method}
%=================================================
%*************************************************
\begin{figure*}[t]
\input{pipeline_figure}
\caption{Illustration of our framework on a toy scan of a table and a chair. We perform geometric partitioning on the  point cloud \Subref{fig:pipea}, which allows us to build the superpoint graph \Subref{fig:pipeb}.
 Each superpoint is embedded by a PointNet network. The embeddings are then refined in GRUs by message passing along superedges to produce the final labeling \Subref{fig:pipec}.}
\label{fig:pipeline}
\end{figure*}
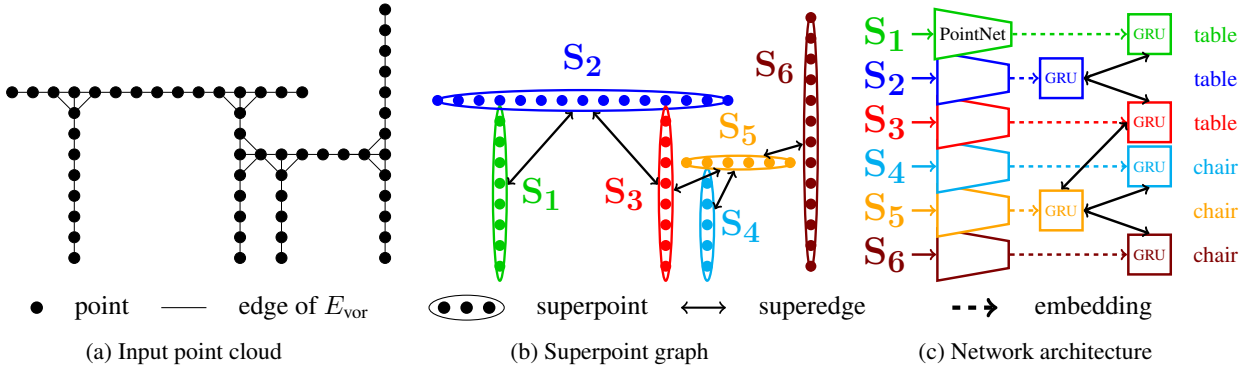
%*************************************************
The main obstacle that our framework tries to overcome is the size of LiDAR scans.
Indeed, they can reach hundreds of millions of points, making direct deep learning approaches intractable. The proposed SPG representation allows us to split the semantic segmentation problem into three distinct problems of different scales, shown in \figref{fig:pipeline}, which can in turn be solved by methods of corresponding complexity:
\begin{itemize}
\item[1] \textbf{Geometrically homogeneous partition:} The first step of our algorithm is to partition the point cloud into geometrically simple yet meaningful shapes, called superpoints. This unsupervised step takes the whole point cloud as input, and therefore must be computationally very efficient. The SPG can be easily computed from this partition.
\item[2] \textbf{Superpoint embedding:} Each node of the SPG corresponds to a small part of the point cloud corresponding to a geometrically simple primitive, which we assume to be semantically homogeneous. Such primitives can be reliably represented by downsampling small point clouds to at most hundreds of points. This small size allows us to utilize recent point cloud embedding methods such as PointNet~\cite{qi2016pointnet}.
\item[3] \textbf{Contextual segmentation:} The graph of superpoints is by orders of magnitude smaller than any graph built on the original point cloud. Deep learning algorithms based on graph convolutions can then be used to classify its nodes using rich edge features facilitating long-range interactions.
\end{itemize}
The SPG representation allows us to perform end-to-end learning of the trainable two last steps. We will describe each step of our pipeline in the following subsections.
%-------------------------------------------------
\subsection{Geometric Partition with a Global Energy}
%-------------------------------------------------   
\label{sec:energy}
In this subsection, we describe our method for partitioning the input point cloud into parts of simple shape.
Our objective is not to retrieve individual objects such as cars or chairs, but rather to break down the objects into simple parts, as seen in \figref{fig:illustration}. However, the clusters being geometrically simple, one can expect them to be semantically homogeneous as well, \ie not to cover objects of different classes. Note that this step of the pipeline is purely unsupervised and makes no use of class labels beyond validation.

We follow the global energy model described by \cite{guinard_weakly_2017} for its computational efficiency. Another advantage is that the segmentation is adaptive to the local geometric complexity. In other words, the segments obtained can be large simple shapes such as roads or walls, as well as much smaller components such as parts of a car or a chair.

Let us consider the input point cloud $C$ as a set of $n$ 3D points. Each point $i\in C$ is defined by its 3D position $p_i$, and, if available, other observations $o_i$ such as color or intensity. For each point, we compute a set of $d_g$ geometric features $f_i\in \bbR^{d_g}$ characterizing the shape of its local neighborhood. In this paper, we use three dimensionality values proposed by \cite{demantke_dimensionality_2011}: linearity, planarity and scattering, as well as the verticality feature introduced by \cite{guinard_weakly_2017}.
We also compute the elevation of each point, defined as the $z$ coordinate of $p_i$ normalized over the whole input cloud.

The global energy proposed by \cite{guinard_weakly_2017} is defined with respect to the $10$-nearest neighbor  \emph{adjacency graph} $\Gnn=\Pa{C,\Enn}$ of the point cloud (note that this is \emph{not} the SPG). The geometrically homogeneous partition is defined as the constant connected components of the solution of the following optimization problem:
\begin{equation}
\label{eq:minimal_partition}
\argmin_{g \in \bbR^{d g}}
\sum_{i \in C}\Norm{g_i - f_i}^2
+ \mu
\sum_{(i,j) \in \Enn}w_{i,j}\Bra{g_i - g_j \neq 0},
\end{equation}
where $\Bra{\cdot}$ is the Iverson bracket. The edge weight $w \in \bbR^{\Abs{E}}_+$ is chosen to be linearly decreasing with respect to the edge length. The factor $\mu$ is the regularization strength and determines the coarseness of the resulting partition.

The problem defined in \eqref{eq:minimal_partition}
is known as \emph{generalized minimal partition problem}, and can be seen as a continuous-space version of the Potts energy model, or an $\ell_0$ variant of the graph total variation.
The minimized functional being nonconvex and noncontinuous implies that the problem cannot realistically be solved exactly for large point clouds. However, the $\ell_0$-cut pursuit algorithm introduced by \cite{landrieu2017cut} is able to quickly find an approximate solution with a few graph-cut iterations. In contrast to other optimization methods such as $\alpha$-expansion \cite{boykov2001fast}, the $\ell_0$-cut pursuit algorithm does not require selecting  the size of the partition in advance.
The constant connected components $\cS=\Cur{S_1, \cdots, S_k}$ of the solution of \eqref{eq:minimal_partition} define our geometrically simple elements, and are referred  as \emph{superpoints} (\ie set of points) in the rest of this paper.
%-------------------------------------------------
\subsection{Superpoint Graph Construction}
%-------------------------------------------------
\label{sec:SPGconstruction}
In this subsection, we describe how we compute the SPG as well as its key features. The SPG is a structured representation of the point cloud, defined as an oriented attributed graph $\cG=\Pa{\cS, \cE, F}$ whose nodes are the set of superpoints $\cS$ and edges $\cE$ (referred to as \emph{superedges}) represent the adjacency between superpoints. The superedges are annotated by a set of $d_f$ features: $F \in \bbR^{\cE \times d_f}$ characterizing the adjacency relationship between superpoints.

We define $\Gvor=\Pa{C,\Evor}$ as the symmetric Voronoi adjacency graph of the complete input point cloud as defined by \cite{jaromczyk1992relative}.
Two superpoints $S$ and $T$ are adjacent if there is at least one edge in $\Evor$ with one end in $S$ and one end in $T$:
\begin{equation}
\label{eq:superedges}
\cE = \Cur{\Pa{S,T} \in \cS^2 \mid \exists \Pa{i,j} \in \Evor \cap \Pa{S \times T}}.
\end{equation}
Important spatial features associated with a superedge $\Pa{S,T}$ are obtained from the set of offsets $\delta(S,T)$ for edges in $\Evor$ linking both superpoints:
\begin{equation} 
\label{eq:delta}
\delta\Pa{S,T}=\Cur{\Pa{p_i-p_j} \mid \Pa{i,j} \in \Evor \cap \Pa{S \times T}}.
\end{equation}
Superedge features can also be derived by comparing the shape and size of the adjacent superpoints. To this end, we compute $\Abs{S}$ as the number of points comprised in a superpoint $S$, as well as shape features $\Length{S}=\lambda_1$, $\Surface{S}=\lambda_1\lambda_2$, $\Volume{S}=\lambda_1\lambda_2\lambda_3$ derived from the eigenvalues $\lambda_1, \lambda_2, \lambda_3$ of the covariance of the positions of the points comprised in each superpoint, sorted by decreasing value. 
In \tabref{table:superedge_features}, we describe a list of the different superedge features used in this paper. Note that the break of symmetry in the edge features makes the SPG a directed graph.
\begin{table}
  \begin{tabular}{|c|c|c|} \hline
  	\small
    Feature name & Size & Description \\\hline
    mean offset & $3$ & $\mathrm{mean}_{m \in \delta\Pa{S,T}}\; \delta_m$\\ 
    offset deviation & $3$ & $\mathrm{std}_{m \in \delta\Pa{S,T}}\; \delta_m$\\ % $\frac1
    centroid offset & $3$ & $\mathrm{mean}_{i \in S}\; p_i - \mathrm{mean}_{j \in T}\; p_j$\\
    length ratio & $1$ & $\log\Length{S} / \Length{T}$\\
    surface ratio & $1$ & $\log\Surface{S} / \Surface{T}$\\
    volume ratio & $1$ & $\log\Volume{S} / \Volume{T}$\\
    point count ratio & $1$ & $\log |S| / |T|$\\
    \hline 
  \end{tabular}
  \caption{List of $d_f=13$ superedge features characterizing the adjacency between two superpoints $S$ and $T$.}
\label{table:superedge_features}
\end{table}
%-------------------------------------------------
\subsection{Superpoint Embedding} \label{sec:embedding}
%-------------------------------------------------
The goal of this stage is to compute a descriptor for every superpoint $S_i$ by embedding it into a vector $\mathbf{z_i}$ of fixed-size dimensionality $d_z$. Note that each superpoint is embedded in isolation; contextual information required for its reliable classification is provided only in the following stage by the means of graph convolutions. 

Several deep learning-based methods have been proposed for this purpose recently. We choose PointNet~\cite{qi2016pointnet} for its remarkable simplicity, efficiency, and robustness. In PointNet, input points are first aligned by a Spatial Transformer Network~\cite{JaderbergSTN}, independently processed by multi-layer perceptrons (MLPs), and finally max-pooled to summarize the shape. 

In our case, input shapes are geometrically simple objects, which can be reliably represented by a small amount of points and embedded by a rather compact PointNet. This is important to limit the memory needed when evaluating many superpoints on current GPUs. In particular, we subsample superpoints on-the-fly down to $n_p=128$ points to maintain efficient computation in batches and facilitate data augmentation. Superpoints of less than $n_p$ points are sampled with replacement, which in principle does not affect the evaluation of PointNet due to its max-pooling. However, we observed that including very small superpoints of less than $n_\mathrm{minp}=40$ points in training harms the overall performance. Thus, embedding of such superpoints is set to zero so that their classification relies solely on contextual information.

In order for PointNet to learn spatial distribution of different shapes, each superpoint is rescaled to unit sphere before embedding. Points are represented by their normalized position $p'_i$, observations $o_i$, and geometric features $f_i$ (since these are already available precomputed from the partitioning step). Furthermore, the original metric diameter of the superpoint is concatenated as an additional feature after PointNet max-pooling in order to stay covariant with shape sizes.
%-------------------------------------------------
\subsection{Contextual Segmentation}
\label{subsec:contexseg}
%-------------------------------------------------
The final stage of the pipeline is to classify each superpoint $S_i$ based on its embedding $\Emb$ and its local surroundings within the SPG. Graph convolutions are naturally suited to this task. In this section, we explain the propagation model of our system.

Our approach builds on the ideas from Gated Graph Neural Networks~\cite{li-ggnn16} and Edge-Conditioned Convolutions (ECC)~\cite{simonovsky2017dynamic}. The general idea is that superpoints refine their embedding according to pieces of information passed along superedges. Concretely, each superpoint $S_i$ maintains its state hidden in a Gated Recurrent Unit (GRU)~\cite{cho-gru14}. The hidden state is initialized with embedding $\Emb$ and is then processed over several iterations (time steps) $t=1\ldots T$. At each iteration $t$, a GRU takes its hidden state $\Gruh{t}{i}$ and an incoming message $\Grum$ as input, and computes its new hidden state $\Gruh{t+1}{i}$. The incoming message $\Grum$ to superpoint $i$ is computed as a weighted sum of hidden states $\Gruh{t}{j}$ of neighboring superpoints $j$. The actual weighting for a superedge $(j,i)$ depends on its attributes $F_{ji,\cdot}$, listed in \tabref{table:superedge_features}. In particular, it is computed from the attributes by a multi-layer perceptron $\Theta$, so-called Filter Generating Network. Formally:
\begin{equation}
\begin{aligned}
\label{eq:gru}
\begin{split}
\Gruh{t+1}{i} &= (1-\Gruu) \odot \Gruq + \Gruu \odot \Gruh{t}{i}\\
\Gruq &= \tanh(\Gruxn{1} + \Grur \odot \Gruhn{1})
\end{split} \\
\begin{split}
\Gruu &= \sigma(\Gruxn{2} + \Gruhn{2}), & \quad \Grur &= \sigma(\Gruxn{3} + \Gruhn{3})
\end{split}
\end{aligned}
\end{equation}
\begin{equation}
\begin{aligned}
\label{eq:gru_norm}
(\Gruhn{1}, \Gruhn{2}, \Gruhn{3})^T &= \rho(W_h \Gruh{t}{i} + b_h) \\
(\Gruxn{1}, \Gruxn{2}, \Gruxn{3})^T &= \rho(W_x \Grux + b_x)
\end{aligned}
\end{equation}
\vspace{-5pt}
\begin{align}
\Grux &= \sigma(W_g \Gruh{t}{i} + b_g) \odot \Grum \label{eq:gru_ig} \\ 
\Grum &= \mathrm{mean}_{j \mid (j,i) \in \cE}\; \Theta(F_{ji,\cdot}; W_e) \odot \Gruh{t}{j} \label{eq:gru_ecc}
\end{align}
\vspace{-24pt}
\begin{align}
\Gruh{1}{i} &= \Emb, & \Gruy &= W_o (\Gruh{1}{i}, \ldots, \Gruh{T+1}{i})^T \label{eq:gru_io},
\end{align}
where $\odot$ is element-wise multiplication, $\sigma(\cdot)$ sigmoid function, and $W_\cdot$ and $b_\cdot$ are trainable parameters shared among all GRUs. \eqref{eq:gru} lists the standard GRU rules~\cite{cho-gru14} with its update gate $\Gruu$ and reset gate $\Grur$. To improve stability during training, in \eqref{eq:gru_norm} we apply Layer Normalization~\cite{BaKH16_LN} defined as $\rho(\mathbf{a}) := (\mathbf{a} - \mathrm{mean}(\mathbf{a})) / (\mathrm{std}(\mathbf{a}) + \epsilon)$ separately to linearly transformed input $\Grux$ and transformed hidden state $\Gruh{t}{i}$, with $\epsilon$ being a small constant. Finally, the model includes three interesting extensions in Equations~\ref{eq:gru_ig}--\ref{eq:gru_io}, which we detail below.

\paragraph*{Input Gating.} We argue that GRU should possess the ability to down-weight (parts of) an input vector based on its hidden state. For example, GRU might learn to ignore its context if its class state is highly certain or to direct its attention to only specific feature channels. \eqref{eq:gru_ig} achieves this by gating message $\Grum$ by the hidden state before using it as input $\Grux$.

\paragraph*{Edge-Conditioned Convolution.} ECC plays a crucial role in our model as it can dynamically generate filtering weights for any value of continuous attributes $F_{ji,\cdot}$ by processing them with a multi-layer perceptron $\Theta$. 
In the original formulation~\cite{simonovsky2017dynamic} (ECC-MV), $\Theta$ regresses a weight matrix to perform matrix-vector multiplication $\Theta(F_{ji,\cdot}; W_e) \Gruh{t}{j}$ for each edge. In this work, we propose a lightweight variant with lower memory requirements and fewer parameters, which is beneficial for datasets with few but large point clouds. Specifically, we regress only an edge-specific weight vector and perform element-wise multiplication as in \eqref{eq:gru_ecc} (ECC-VV). Channel mixing, albeit in an edge-unspecific fashion, is postponed to \eqref{eq:gru_norm}. Finally, let us remark that $\Theta$ is shared over time iterations and that self-loops as proposed in~\cite{simonovsky2017dynamic} are not necessary due to the existence of hidden states in GRUs.

\paragraph*{State Concatenation.} Inspired by DenseNet~\cite{densenet17}, we concatenate hidden states over all time steps and linearly transform them to produce segmentation logits $\Gruy$ in \eqref{eq:gru_io}. This allows to exploit the dynamics of hidden states due to increasing receptive field for the final classification. 

\paragraph*{Relation to CRFs.} In image segmentation, post-processing of convolutional outputs using Conditional Random Fields (CRFs) is widely popular. Several inference algorithms can be formulated as (recurrent) network layers amendable to end-to-end learning~\cite{Zheng15crf,SchwingU15}, possibly with general pairwise potentials \cite{LinSHR16,ChandraK16,LarssonK0ATH17}. While our method of information propagation shares both these characteristics, our GRUs operate on $d_z$-dimensional intermediate feature space, which is richer and less constrained than low-dimensional vectors representing beliefs over classes, as also discussed in~\cite{gadde16bi}. Such enhanced access to information is motivated by the desire to learn a powerful representation of context, which goes beyond belief compatibilities, as well as the desire to be able to discriminate our often relatively weak unaries (superpixel embeddings). We empirically evaluate these claims in \secref{subsec:ablation}.

%-------------------------------------------------    
\subsection{Further Details}
%-------------------------------------------------
\label{sec:details}

\paragraph*{Adjacency Graphs.}
In this paper, we use two different adjacency graphs between points of the input clouds: $\Gnn$ in \secref{sec:energy} and $\Gvor$ in \secref{sec:SPGconstruction}.
Indeed, different definitions of adjacency have different advantages. Voronoi adjacency is more suited to capture long-range relationships between superpoints, which is beneficial for the SPG.
Nearest neighbors adjacency tends not to connect objects separated by a small gap. This is desirable for the global energy but tends to produce a SPG with many small connected components, decreasing embedding quality.
Fixed radius adjacency should be avoided in general as it handles the variable density of LiDAR scans poorly.

\paragraph*{Training.} While the geometric partitioning step is unsupervised, superpoint embedding and contextual segmentation are trained jointly in a supervised way with cross entropy loss. Superpoints are assumed to be semantically homogeneous and, consequently, assigned a hard ground truth label corresponding to the majority label among their contained points. We also considered using soft labels corresponding to normalized histograms of point labels and training with Kullback-Leibler~\cite{Kullback_and_Leibler_1951} divergence loss. It performed slightly worse in our initial experiments, though.

Naive training on large SPGs may approach memory limits of current GPUs. We circumvent this issue by randomly subsampling the sets of superpoints at each iteration and training on induced subgraphs, \ie graphs composed of subsets of nodes and the original edges connecting them. Specifically, graph neighborhoods of order $3$ are sampled to select at most $512$ superpoints per SPG with more than $n_\mathrm{minp}$ points, as smaller superpoints are not embedded. Note that as the induced graph is a union of small neighborhoods, relationships over many hops may still be formed and learned. This strategy also doubles as data augmentation and a strong regularization, together with randomized sampling of point clouds described in \secref{sec:embedding}. Additional data augmentation is performed by randomly rotating superpoints around the vertical axis and jittering point features by Gaussian noise $\cN(0,0.01)$ truncated to $[-0.05,0.05]$.

\paragraph*{Testing.} In modern deep learning frameworks, testing can be made very memory-efficient by discarding layer activations as soon as the follow-up layers have been computed. In practice, we were able to label full SPGs at once. To compensate for randomness due to subsampling of point clouds in PointNets, we average logits obtained over $10$ runs with different seeds.
%=================================================
\section{Experiments}
%=================================================
We evaluate our pipeline on the two currently largest point cloud segmentation benchmarks, Semantic3D \cite{hackel2017semantic3d} and Stanford Large-Scale 3D Indoor Spaces (S3DIS) \cite{armeni_cvpr16}, on both of which we set the new state of the art. Furthermore, we perform an ablation study of our pipeline in \secref{subsec:ablation}.

Even though the two data sets are quite different in nature (large outdoor scenes for Semantic3D, smaller indoor scanning for S3DIS), we use nearly the same model for both. The deep model is rather compact and $6$ GB of GPU memory is enough for both testing and training. We refer to \ARXIV{Appendix~\ref{sec:model_details}}{the Supplementary} for precise details on hyperparameter selection, architecture configuration, and training procedure.

Performance is evaluated using three metrics: per-class intersection over union (IoU), per-class accuracy (Acc), and overall accuracy (OA), defined as the proportion of correctly classified points. We stress that the metrics are computed on the original point clouds, not on superpoints. 
%-------------------------------------------------
\subsection{Semantic3D}
%-------------------------------------------------
Semantic3D \cite{hackel2017semantic3d} is the largest available LiDAR  dataset with over 3 billion points from a variety of urban and rural scenes. Each point has RGB and intensity values (the latter of which we do not use). The dataset consists of 15 training scans and 15 test scans with withheld labels. We also evaluate on the reduced set of 4 subsampled scans, as common in past work.

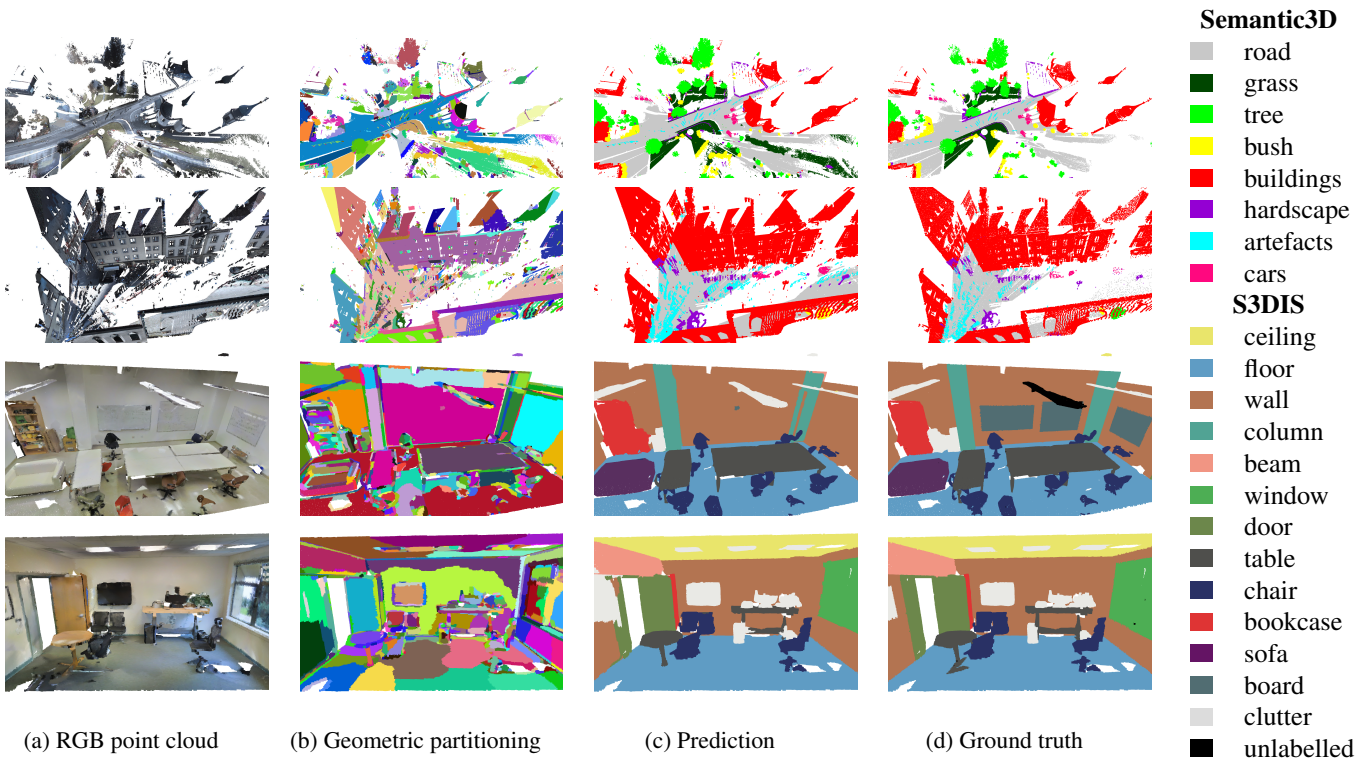
\begin{figure*}
\input{illustration}
\caption{Example visualizations on both datasets. The colors in \Subref{fig:illustration_seg} are chosen randomly for each element of the partition.}
\label{fig:illustration}
\end{figure*}

In \tabref{tab:results_semantic3D}, we provide the results of our algorithm compared to other state of the art recent algorithms and in \figref{fig:illustration}, we provide qualitative results of our framework.
Our framework improves significantly on the state of the art of semantic segmentation for this data set, \ie by nearly 12 mIoU points on the reduced set and by nearly 9 mIoU points on the full set. In particular, we observe a steep gain on the "artefact" class. This can be explained by the ability of the partitioning algorithm to detect artifacts due to their singular shape, while they are hard to capture using snapshots, as suggested by \cite{boulch2017unstructured}. Furthermore, these small object are often merged with the road when performing spatial regularization. 
%**********************************
\begin{table*}\begin{center}\small
\begin{tabular}{|c|C{0.06\textwidth}C{0.06\textwidth}|C{0.06\textwidth}C{0.06\textwidth}C{0.06\textwidth}C{0.06\textwidth}C{0.06\textwidth}C{0.06\textwidth}C{0.06\textwidth}C{0.06\textwidth}|}\hline
\bf Method& OA & mIoU&
\begin{tabular}{@{}c@{}} man-made\\ terrain\end{tabular}&
\begin{tabular}{@{}c@{}} natural\\ terrain\end{tabular}&
\begin{tabular}{@{}c@{}} high\\ vegetation\end{tabular}&
\begin{tabular}{@{}c@{}} low \\ vegetation\end{tabular}&
buildings&
\begin{tabular}{@{}c@{}} hard-\\ scape\end{tabular}&
\begin{tabular}{@{}c@{}} scanning\\ artefact\end{tabular}&
cars
\\\hline
\multicolumn{3}{|c}{reduced test set: $\num{78 699 329}$ points}&
\multicolumn{8}{c|}{}
\\\hline
TMLC-MSR \cite{hackel2016fast} &86.2&54.2&89.8&74.5&53.7&26.8&88.8&18.9&36.4&44.7\\
DeePr3SS \cite{lawin2017deep}
&88.9&58.5&85.6&83.2&74.2&32.4&89.7&18.5&25.1&59.2\\ 
SnapNet \cite{boulch2017unstructured}
&88.6&59.1&82.0&77.3&79.7&22.9&91.1&18.4&37.3&64.4\\
SegCloud \cite{tchapmi2017segcloud}
&88.1&61.3&83.9&66.0&86.0&40.5&91.1&30.9&27.5&64.3\\
SPG (Ours)&\bf 94.0&\bf73.2&\bf97.4&\bf92.6&\bf87.9&\bf44.0&\bf93.2&\bf31.0&\bf63.5&\bf76.2\\\hline
\multicolumn{3}{|c}{full test set: $\num{2 091 952 018}$ points}&
\multicolumn{8}{c|}{}
\\\hline 
TMLC-MS \cite{hackel2016fast}
&85.0&49.4&91.1&69.5&32.8&21.6&87.6&25.9&11.3&55.3\\
SnapNet \cite{boulch2017unstructured}
&91.0&67.4&89.6&\bf79.5&74.8&56.1&90.9&36.5&34.3&77.2\\
SPG (Ours)&\bf92.9&\bf76.2&\bf91.5&75.6&\bf78.3&\bf71.7&\bf94.4&\bf56.8 &\bf52.9&\bf88.4\\\hline
\end{tabular}
\end{center}
\caption{Intersection over union metric for the different classes of the Semantic3D dataset. OA is the global accuracy, while mIoU refers to the unweighted average of IoU of each class.}
\label{tab:results_semantic3D}
\end{table*}

%**********************************
%**********************************
%-------------------------------------------------
\subsection{Stanford Large-Scale 3D Indoor Spaces}
%-------------------------------------------------
The S3DIS dataset \cite{armeni_cvpr16} consists of 3D RGB point clouds of six floors from three different buildings split into individual rooms.
We evaluate our framework following two dominant strategies found in previous works. As advocated by~\cite{qi2016pointnet,Engelmann17_3dsemseg}, we perform $6$-fold cross validation with micro-averaging, \ie computing metrics once over the merged predictions of all test folds. Following \cite{tchapmi2017segcloud}, we also report the performance on the fifth fold only (Area 5), corresponding to a building not present in the other folds.
Since some classes in this data set cannot be partitioned purely using geometric features (such as boards or paintings on walls), we concatenate the color information $o$ to the geometric features $f$ for the partitioning step.

The quantitative results are displayed in \tabref{tab:results_S3DIS}, with qualitative results in \figref{fig:illustration} and in \ARXIV{Appendix~\ref{sec:video}}{the Supplementary}.
S3DIS is a difficult dataset with hard to retrieve classes such as white boards on white walls and columns within walls. From the quantitative results we can see that our framework performs better than other methods on average. Notably, doors are able to be correctly classified at a higher rate than other approaches, as long as they are open, as illustrated in \figref{fig:illustration}. Indeed, doors are geometrically similar to walls, but their position with respect to the door frame allows our network to retrieve them correctly. On the other hand, the partition merges white boards with walls, depriving the network from the opportunity to even learn to classify them: the IoU of boards for theoretical perfect classification of superpoints (as in \secref{subsec:ablation}) is only $51.3$.

\textbf{Computation Time.} In \tabref{tab:computation_time}, we report computation time over the different steps of our pipeline for the inference on Area 5 measured on a 4 GHz CPU and GTX 1080 Ti GPU. While the bulk of time is spent on the CPU for partitioning and SPG computation, we show that voxelization as pre-processing, detailed in \ARXIV{Appendix~\ref{sec:model_details}}{Supplementary}, leads to a significant speed-up as well as improved accuracy.

%**********************************
\begin{table*}\begin{center}
\resizebox{1\linewidth}{!}{
\begin{tabular}{|c|C{0.04\textwidth}C{0.04\textwidth}C{0.04\textwidth}|C{0.04\textwidth}C{0.04\textwidth}C{0.04\textwidth}C{0.04\textwidth}C{0.05\textwidth}C{0.05\textwidth}C{0.04\textwidth}C{0.04\textwidth}C{0.04\textwidth}C{0.06\textwidth}C{0.04\textwidth}C{0.04\textwidth}C{0.04\textwidth}|}\hline
\bf Method & OA & mAcc & mIoU & ceiling & floor & wall & beam & column & window & door & chair & table & bookcase & sofa & board & clutter\\\hline
A5 PointNet \cite{qi2016pointnet}&--&48.98&41.09&88.80&\bf97.33&69.80&\bf0.05 &3.92&46.26&10.76&52.61&58.93&40.28&5.85&\bf26.38&33.22\\ 
A5 SEGCloud \cite{tchapmi2017segcloud}&--&57.35&48.92&\bf90.06 &96.05 &69.86&0.00&18.37&38.35&23.12&75.89&70.40&58.42&40.88 &12.96&41.60\\
A5 SPG (Ours)
&86.38&\bf66.50&\bf58.04&
89.35&96.87&\bf78.12&0.0&\bf42.81
&\bf48.93&\bf61.58&\bf84.66&\bf75.41&
\bf69.84&\bf52.60&2.10&\bf52.22\\
\hline\hline
PointNet \cite{qi2016pointnet} in \cite{Engelmann17_3dsemseg}&78.5&66.2&47.6&88.0&88.7&69.3&42.4&23.1&47.5&51.6&42.0&54.1&38.2&9.6&29.4&35.2\\
Engelmann \etal~\cite{Engelmann17_3dsemseg}&81.1&66.4&49.7&\bf90.3&92.1&67.9&44.7&24.2&52.3&51.2&47.4&58.1&39.0&6.9&\bf30.0&41.9\\
SPG (Ours)&\bf85.5&\bf73.0&\bf62.1&89.9&\bf95.1&\bf76.4&\bf62.8&\bf47.1&\bf55.3&\bf68.4&\bf73.5&\bf69.2&\bf63.2&\bf45.9&8.7&\bf52.9\\\hline
\end{tabular}}
\end{center}
\caption{Results on the S3DIS dataset on fold ``Area 5'' (top) and micro-averaged over all 6 folds (bottom). Intersection over union is shown split per class.}
\label{tab:results_S3DIS}
\end{table*}
%**********************************
\begin{table}\begin{center}
\resizebox{1\linewidth}{!}{
\begin{tabular}{|c|cccc|}\hline
Step & Full cloud & $2$ cm & $3$ cm & $4$ cm\\\hline
Voxelization & $\num{0}$ &$\num{40}$&$\num{24}$&$\num{16}$\\
Feature computation & $\num{439}$&$\num{194}$&$\num{88}$&$\num{43}$\\
Geometric partition & $\num{3 428}$&$\num{1013}$&$\num{447}$&$\num{238}$\\
SPG computation & $\num{3 800}$&$\num{958}$&$\num{436}$&$\num{252}$\\
Inference & $10 \times 24$ & $10 \times 11$ & $10 \times 6$ & $10 \times 5$\\\hline
Total &$\num{7 907}$&$\num{2315}$&$\num{1055}$&$\num{599}$\\
mIoU $6$-fold&54.1&60.2&62.1&57.1\\\hline
\end{tabular}}
\end{center}
\caption{Computation time in seconds for the inference on S3DIS Area 5 ($68$ rooms, $\num{78 649 682}$ points) for different voxel sizes.}
\label{tab:computation_time}
\end{table}
%**********************************
%-------------------------------------------------
\subsection{Ablation Studies} \label{subsec:ablation}
%------------------------------------------------
To better understand the influence of various design choices made in our framework, we compare it to several baselines and perform an ablation study. Due to the lack of public ground truth for test sets of Semantic3D, we evaluate on S3DIS with 6-fold cross validation and show comparison of different models to our $\mathrm{Best}$ model in \tabref{tab:results_ablation}.

\textbf{Performance Limits.} The contribution of contextual segmentation can be bounded both from below and above. The lower bound ($\mathrm{Unary}$) is estimated by training PointNet with $d_z=13$ but otherwise the same architecture, denoted as PointNet13, to directly predict class logits, without SPG and GRUs. The upper bound ($\mathrm{Perfect}$) corresponds to assigning each superpoint its ground truth label, and thus sets the limit of performance due to the geometric partition. We can see that contextual segmentation is able to win roughly $22$ mIoU points over unaries, confirming its importance. Nevertheless, the learned model still has room of up to $26$ mIoU points for improvement, while about $12$ mIoU points are forfeited to the semantic inhomogeneity of superpoints.

\textbf{CRFs.} We compare the effect of our GRU+ECC-based network to CRF-based regularization. As a baseline ($\mathrm{iCRF}$), we post-process $\mathrm{Unary}$ outputs by CRF inference over SPG connectivity with scalar transition matrix, as described by \cite{guinard_weakly_2017}. Next ($\mathrm{CRF-ECC}$), we adapt CRF-RNN framework of Zheng \etal~\cite{Zheng15crf} to general graphs with edge-conditioned convolutions (see \ARXIV{Appendix~\ref{sec:crfecc}}{Supplementary} for details) and train it with PointNet13 end-to-end. Finally ($\mathrm{GRU13}$), we modify $\mathrm{Best}$ to use PointNet13. We observe that $\mathrm{iCRF}$ barely improves accuracy (+1 mIoU), which is to be expected, since the partitioning step already encourages spatial regularity. $\mathrm{CRF-ECC}$ does better (+15 mIoU) due to end-to-end learning and use of edge attributes, though it is still below $\mathrm{GRU13}$ (+18 mIoU), which performs more complex operations and does not enforce normalization of the embedding. Nevertheless, the 32 channels used in $\mathrm{Best}$ instead of the 13 used in $\mathrm{GRU13}$ provide even more freedom for feature representation (+22 mIoU).

\textbf{Ablation.} We explore the advantages of several design choices by individually removing them from $\mathrm{Best}$ in order to compare the framework's performance with and without them. In $\mathrm{NoInputGate}$ we remove input gating in GRU; in $\mathrm{NoConcat}$ we only consider the last hidden state in GRU for output as $\Gruy = W_o \Gruh{T+1}{i}$ instead of concatenation of all steps;
in $\mathrm{NoEdgeFeat}$ we perform homogeneous regularization by setting all superedge features to scalar $1$; and in $\mathrm{ECC-VV}$ we use the proposed lightweight formulation of ECC.
We can see that each of the first two choices accounts for about $5$ mIoU points. Next, without edge features our method falls back even below $\mathrm{iCRF}$ to the level of $\mathrm{Unary}$, which validates their design and overall motivation for SPG. $\mathrm{ECC-VV}$ decreases the performance  on the S3DIS dataset by $3$ mIoU points, whereas it has improved the performance on Semantic3D by $2$ mIoU. Finally, we invite the reader to \ARXIV{Appendix~\ref{sec:ext_ablation}}{Supplementary} for further ablations.

\begin{table}\begin{center}\small
\begin{tabular}{|c|cc|}\hline
Model & mAcc & mIoU \\\hline
$\mathrm{Best}$ & 73.0 & 62.1  \\
$\mathrm{Perfect}$ & 92.7 & 88.2\\
$\mathrm{Unary}$ & 50.8 & 40.0 \\
$\mathrm{iCRF}$ & 51.5 &  40.7 \\
$\mathrm{CRF-ECC}$ & 65.6 & 55.3\\
$\mathrm{GRU13}$ & 69.1 & 58.5 \\
$\mathrm{NoInputGate}$ & 68.6 & 57.5  \\
$\mathrm{NoConcat}$ & 69.3 & 57.7 \\
$\mathrm{NoEdgeFeat}$ & 50.1 & 39.9\\
$\mathrm{ECC-VV}$ & 70.2& 59.4 \\
\hline
\end{tabular}
\end{center}
\caption{Ablation study and comparison to various baselines on S3DIS (6-fold cross validation).}
\label{tab:results_ablation}
\end{table}

%=================================================
\section{Conclusion}
%=================================================
We presented a deep learning framework for performing semantic segmentation of large point clouds based on a partition into simple shapes. We showed that SPGs allow us to use effective deep learning tools, which would not be able to handle the data volume otherwise. Our method significantly improves on the state of the art on two publicly available datasets. Our experimental analysis suggested that future improvements can be made in both partitioning and learning deep contextual classifiers.

The source code in PyTorch as well as the trained models are available at \url{https://github.com/loicland/superpoint_graph}.

{\small
\bibliographystyle{ieee}
\bibliography{egbib}
}

\ARXIV{
\appendix
\section*{Appendix} 
\input{supplementary}
}{}

\end{document}

%% file: pipeline_figure.tex
\begin{tabular}[b]{ccc}
\begin{subfigure}[b]{0.3\textwidth}
\begin{tabular}[b]{c}
\resizebox{1\textwidth}{!}{
\begin{tikzpicture}
\tikzstyle{point}=[circle, draw = black, fill = black, minimum size = 0.5mm, scale = 0.8]
\tikzstyle{point_table_top}      =[point]
\tikzstyle{point_table_left_leg} =[point]
\tikzstyle{point_table_right_leg}=[point]
\tikzstyle{point_chair_left_leg} =[point]
\tikzstyle{point_chair_right_leg}=[point]
\tikzstyle{point_chair_seat}     =[point]
\tikzstyle{point_chair_back}     =[point]
\tikzstyle{cluster}=[ellipse, ultra thick, fill=none]
\tikzstyle{edge}=[-]
%---ghost ellipses-------
%needed for perfect alignment
%tabletop
\node[cluster, draw = white, minimum height =4.6cm, minimum width = 0.25cm, label={[right=2mm, white]0:\Huge$\mathbf{S_1}$}] at (1.5,1.75)  (cluster1) {};
\node[cluster, draw = white,  minimum height =0.5cm, minimum width = 7cm, label={[above=2mm, white]90:\Huge$\mathbf{S_2}$}] at (3.5,4)  (cluster2) {};
\node[cluster, draw = white, minimum height =4.2cm, minimum width = 0.25cm, label={[left=2mm, white]180:\Huge$\mathbf{S_3}$}] at (5.5,1.75)  (cluster3) {};
\node[cluster, draw = white, minimum height =2.7cm, minimum width = 0.25cm, label={[right=0mm, white]350:\Huge$\mathbf{S_4}$}] at (6.5,1)  (cluster4) {};
\node[cluster, draw = white, fill=none, minimum height =0.25cm, minimum width = 2.7cm, label={[above=1mm, white]90:\Huge$\mathbf{S_5}$}] at (7.25,2.5)  (cluster5) {};
\node[cluster, draw = white, fill=none, minimum height =6.6cm, minimum width = 0.25cm, label={[left=1mm, white]95:\Huge$\mathbf{S_6}$}] at (9,3)  (cluster6) {};
%---point cloud----
\input{table_and_chair}
%---edges----------
%--tabletop--------
\draw [edge] (table_top1) -- (table_top2);
\draw [edge] (table_top2) -- (table_top3);
\draw [edge] (table_top3) -- (table_top4);
\draw [edge] (table_top4) -- (table_top5);
\draw [edge] (table_top5) -- (table_top6);
\draw [edge] (table_top6) -- (table_top7);
\draw [edge] (table_top7) -- (table_top8);
\draw [edge] (table_top8) -- (table_top9);
\draw [edge] (table_top9) -- (table_top10);
\draw [edge] (table_top10) -- (table_top11);
\draw [edge] (table_top11) -- (table_top12);
\draw [edge] (table_top12) -- (table_top13);
\draw [edge] (table_top13) -- (table_top14);
\draw [edge] (table_top14) -- (table_top15);
%--tabletop to legs--------
\draw [edge] (table_top3)  -- (table_left_leg8);
\draw [edge] (table_top4)  -- (table_left_leg8);
\draw [edge] (table_top5)  -- (table_left_leg8);
\draw [edge] (table_top11) -- (table_right_leg8);
\draw [edge] (table_top12) -- (table_right_leg8);
\draw [edge] (table_top13) -- (table_right_leg8);
%---table_legs---------------
\draw [edge] (table_right_leg1) -- (table_right_leg2);
\draw [edge] (table_right_leg2) -- (table_right_leg3);
\draw [edge] (table_right_leg3) -- (table_right_leg4);
\draw [edge] (table_right_leg4) -- (table_right_leg5);
\draw [edge] (table_right_leg5) -- (table_right_leg6);
\draw [edge] (table_right_leg6) -- (table_right_leg7);
\draw [edge] (table_right_leg7) -- (table_right_leg8);
\draw [edge] (table_left_leg1)  -- (table_left_leg2);
\draw [edge] (table_left_leg2)  -- (table_left_leg3);
\draw [edge] (table_left_leg3)  -- (table_left_leg4);
\draw [edge] (table_left_leg4)  -- (table_left_leg5);
\draw [edge] (table_left_leg5)  -- (table_left_leg6);
\draw [edge] (table_left_leg6)  -- (table_left_leg7);
\draw [edge] (table_left_leg7)  -- (table_left_leg8);
%---chair_legs--------------------
\draw [edge] (chair_right_leg1) -- (chair_right_leg2);
\draw [edge] (chair_right_leg2) -- (chair_right_leg3);
\draw [edge] (chair_right_leg3) -- (chair_right_leg4);
\draw [edge] (chair_right_leg4) -- (chair_right_leg5);
\draw [edge] (chair_left_leg1)  -- (chair_left_leg2);
\draw [edge] (chair_left_leg2)  -- (chair_left_leg3);
\draw [edge] (chair_left_leg3)  -- (chair_left_leg4);
\draw [edge] (chair_left_leg4)  -- (chair_left_leg5);
%---chair_seat--------------------
\draw [edge] (chair_seat1) -- (chair_seat2);
\draw [edge] (chair_seat2) -- (chair_seat3);
\draw [edge] (chair_seat3) -- (chair_seat4);
\draw [edge] (chair_seat4) -- (chair_seat5);
\draw [edge] (chair_seat5) -- (chair_seat6);
\draw [edge] (chair_seat5) -- (chair_seat7);
%---chair_back--------------------
\draw [edge] (chair_back1) -- (chair_back2);
\draw [edge] (chair_back2) -- (chair_back3);
\draw [edge] (chair_back3) -- (chair_back4);
\draw [edge] (chair_back4) -- (chair_back5);
\draw [edge] (chair_back5) -- (chair_back6);
\draw [edge] (chair_back6) -- (chair_back7);
%---internal chair edges----------
\draw [edge] (chair_left_leg5)  -- (chair_seat1);
\draw [edge] (chair_left_leg5)  -- (chair_seat2);
\draw [edge] (chair_left_leg5)  -- (chair_seat3);
\draw [edge] (chair_right_leg5) -- (chair_seat6);
\draw [edge] (chair_right_leg5) -- (chair_seat7);
\draw [edge] (chair_back1)      -- (chair_seat6);
\draw [edge] (chair_back1)      -- (chair_seat7);
%---edge to chair edges----------
\draw [edge] (chair_seat1)  -- (table_right_leg5);
\draw [edge] (chair_seat1)  -- (table_right_leg6);
\draw [edge] (chair_seat1)  -- (table_right_leg7);
\end{tikzpicture}
}\\
\begin{tabular}[b]{cccc}
\begin{tikzpicture}[baseline=-0.5ex]
\node at (0,0) [circle, draw = none, fill, minimum size = 0.5mm, scale = 0.5]   (toy_point){};
\end{tikzpicture}
&
point
&
\begin{tikzpicture}[baseline=-0.5ex]
\draw [-, draw = black] (0,0) -- (.6,0);
\end{tikzpicture}
&
edge of $\Evor$
\end{tabular}
\end{tabular}
\caption{Input point cloud}
\label{fig:pipea}
\end{subfigure}
&
\begin{subfigure}[b]{0.3\textwidth}
\begin{tabular}[b]{c}
\resizebox{1\textwidth}{!}{
\begin{tikzpicture}
\tikzstyle{point}=[circle, draw = none, fill, minimum size = 0.5mm, scale = 0.8]
\tikzstyle{point_table_top}      =[point, fill = blue ]
\tikzstyle{point_table_left_leg} =[point, fill = green!80!black]
\tikzstyle{point_table_right_leg}=[point, fill = red]
\tikzstyle{point_chair_left_leg} =[point, fill = cyan]
\tikzstyle{point_chair_right_leg}=[point, fill = Maroon]
\tikzstyle{point_chair_seat}     =[point, fill = Orange]
\tikzstyle{point_chair_back}     =[point, fill = Maroon]
\tikzstyle{cluster}=[ellipse, ultra thick, fill=none]
\tikzstyle{superedge}=[<->, ultra thick]
%---point cloud----
\input{table_and_chair}
%---ellipses-------
\node[cluster, draw = green!80!black, minimum height =4.2cm, minimum width = 0.25cm, label={[right=2mm, green!80!black]0:\Huge$\mathbf{S_1}$}] at (1.5,1.75)  (cluster1) {};
\node[cluster, draw = blue,  minimum height =0.5cm, minimum width = 7cm, label={[above=2mm, blue]90:\Huge$\mathbf{S_2}$}] at (3.5,4)  (cluster2) {};
\node[cluster, draw = red, minimum height =4.2cm, minimum width = 0.25cm, label={[left=2mm, red]180:\Huge$\mathbf{S_3}$}] at (5.5,1.75)  (cluster3) {};
\node[cluster, draw = cyan, minimum height =2.7cm, minimum width = 0.25cm, label={[right=0mm, cyan]350:\Huge$\mathbf{S_4}$}] at (6.5,1)  (cluster4) {};
\node[cluster, draw = Orange, fill=none, minimum height =0.25cm, minimum width = 2.7cm, label={[above=1mm, Orange]90:\Huge$\mathbf{S_5}$}] at (7.25,2.5)  (cluster5) {};
\node[cluster, draw = Maroon, fill=none, minimum height =6.2cm, minimum width = 0.25cm, label={[left=1mm, Maroon]95:\Huge$\mathbf{S_6}$}] at (9,3)  (cluster6) {};
%---superedges-----------
\draw [superedge] (cluster1)  to  (cluster2);
\draw [superedge] (cluster2)  to  (cluster3);
\draw [superedge] (cluster3)  to  (cluster5);
\draw [superedge] (cluster4)  to  (cluster5);
\draw [superedge] (cluster5)  to  (cluster6); 
\end{tikzpicture}
}
\end{tabular}\\
\begin{tabular}[b]{cccc}
\begin{tikzpicture}[baseline=-0.5ex]
\node at (0.2,0) [circle, draw = none, fill, minimum size = 0.5mm, scale = 0.5]   (toy_point1){};
\node at (0.5,0) [circle, draw = none, fill, minimum size = 0.5mm, scale = 0.5]   (toy_point2){};
\node at (0.8,0) [circle, draw = none, fill, minimum size = 0.5mm, scale = 0.5]   (toy_point3){};
\node[ellipse, draw = black, minimum height =0.15cm, minimum width = 1cm] at (0.5,0)  (cluster1) {};
\end{tikzpicture}
&
superpoint
&
\begin{tikzpicture}[baseline=-0.5ex]
\draw [<->, thick, draw = black] (0,0) -- (.6,0);
\end{tikzpicture}
&
superedge
\end{tabular}
\caption{Superpoint graph}
\label{fig:pipeb}
\end{subfigure}
&
\begin{subfigure}[b]{0.3\textwidth}
\begin{tabular}[b]{c}
\resizebox{1\textwidth}{!}{
\begin{tikzpicture}
\tikzstyle{superpointname}=[draw = none, fill = none]
\tikzstyle{pointnet}=[trapezium, trapezium angle=80, draw, minimum height=.25cm, minimum width=.5cm, shape border rotate=270, ultra thick,fill=white]
\tikzstyle{arrow}=[->, ultra thick]
\tikzstyle{superpoint}=[rectangle, fill = none, minimum size = .9cm, ultra thick]
\tikzstyle{embedding}=[->, ultra thick, dashed]
\tikzstyle{superedge}=[<->, ultra thick]
\tikzstyle{result}=[rectangle, fill = none, minimum size = 1cm, ultra thick]
 %---superpointnames-----
 \node[superpointname] at (0,5)  [text = green!80!black  ]  (superpointname1){\Huge$\mathbf{S_1}$};
  \node[superpointname] at (0,4)  [text = blue   ]  (superpointname2){\Huge$\mathbf{S_2}$};
  \node[superpointname] at (0,3)  [text = red    ]  (superpointname3){\Huge$\mathbf{S_3}$};
  \node[superpointname] at (0,2)  [text = cyan   ]  (superpointname4){\Huge$\mathbf{S_4}$};
  \node[superpointname] at (0,1) [text = Orange ]  (superpointname6){\Huge$\mathbf{S_5}$};
   \node[superpointname] at (0,0)  [text = Maroon   ]  (superpointname5){\Huge$\mathbf{S_6}$};
  %---pointnets-----------
\node[pointnet] at (2,0)  [draw = Maroon   , text = white]  (pointnet5){\large pointnet};
\node[pointnet] at (2,1)  [draw = Orange , text = white]  (pointnet6){\large pointnet};
\node[pointnet] at (2,2)  [draw = cyan   , text = white]  (pointnet4){\large pointnet};
\node[pointnet] at (2,3)  [draw = red    , text = white]  (pointnet3){\large pointnet};
\node[pointnet] at (2,4)  [draw = blue   , text = white]  (pointnet2){\large pointnet};
\node[pointnet] at (2,5)  [draw = green!80!black  , text = black]  (pointnet1){\large PointNet};
 %---arrows-----------
  \draw [arrow, draw = green!80!black  ] (superpointname1)  -- (pointnet1);
  \draw [arrow, draw = blue ] (superpointname2)  -- (pointnet2);
  \draw [arrow, draw = red   ] (superpointname3)  -- (pointnet3);
  \draw [arrow, draw = cyan  ] (superpointname4)  -- (pointnet4); 	
  \draw [arrow, draw = Maroon  ] (superpointname5)  -- (pointnet5);
  \draw [arrow, draw = Orange] (superpointname6)  -- (pointnet6);
  %---superpoints-------
  \node[superpoint] at (4  ,4 )  [draw = blue   , text = blue  ]  (GRU2){GRU};
  \node[superpoint] at (6  ,5)  [draw = green!80!black  , text = green!80!black ]  (GRU1){GRU};
  \node[superpoint] at (6  ,3 )  [draw = red    , text = red   ]  (GRU3){GRU};
  \node[superpoint] at (6  ,2 )  [draw = cyan   , text = cyan  ]  (GRU4){GRU};
  \node[superpoint] at (6  ,0 )  [draw = Maroon   , text = Maroon  ]  (GRU5){GRU};
  \node[superpoint] at (4  ,1 )  [draw = Orange , text =Orange ]  (GRU6){GRU};
  %---embedding---------
  \draw [embedding, draw = green!80!black ] (pointnet1)  to (GRU1);
  \draw [embedding, draw = blue ] (pointnet2) to (GRU2);
  \draw [embedding, draw = red   ] (pointnet3) to  (GRU3);
  \draw [embedding, draw = cyan  ] (pointnet4) to  (GRU4); 	
  \draw [embedding, draw = Maroon  ] (pointnet5) to  (GRU5);
  \draw [embedding, draw = Orange] (pointnet6) to  (GRU6);
  %---superedges-----------
\draw [superedge] (GRU1.south) to (GRU2.east);
\draw [superedge] (GRU2.east)  to (GRU3.north);
\draw [superedge] (GRU3.west)  to (GRU6.north);
\draw [superedge] (GRU4.south) to (GRU6.east) ;
\draw [superedge] (GRU5.north) to (GRU6.east) ;
 %---result-----
 \node[result] at (7.5,5)  [text = green!80!black , draw = white]  (result1){\Large table};
 \node[result] at (7.5,4)  [text = blue  , draw = white]  (result2){\Large table};
 \node[result] at (7.5,3)  [text = red   , draw = white]  (result3){\Large table};
 \node[result] at (7.5,2)  [text = cyan  , draw = white]  (result4){\Large chair};
 \node[result] at (7.5,1)  [text = Orange, draw = white]  (result5){\Large chair};
 \node[result] at (7.5,0)  [text = Maroon, draw = white]  (result6){\Large chair};
\end{tikzpicture}
}\\ 
\begin{tikzpicture}[baseline=-0.5ex]
\draw [->, ultra thick, draw = black, dashed] (0,0) -- (.6,0);
\end{tikzpicture}
\;\;\;
embedding
\end{tabular}
\caption{Network architecture}
\label{fig:pipec}
\end{subfigure}
\end{tabular}

%% file: illustration.tex
%\begin{tabular}[b]{ccccc}
\begin{tabular}[b]{ccccc}
 \begin{subfigure}[b]{0.2\textwidth}
  \parbox[b]{1\textwidth}{
 \begin{tabular}[b]{c}
 	\includegraphics[width=1\textwidth]{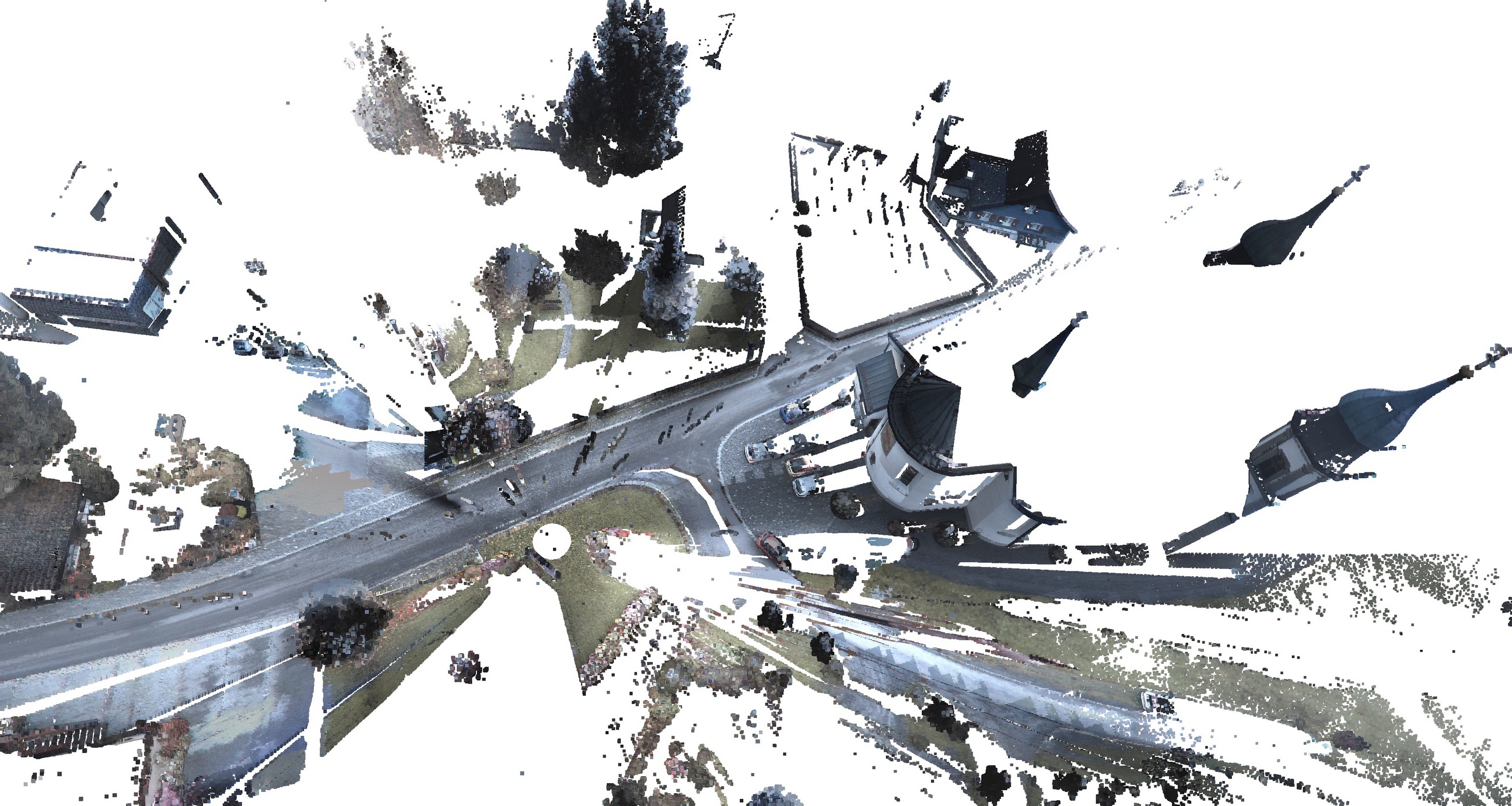}\\
    \includegraphics[width=1\textwidth] {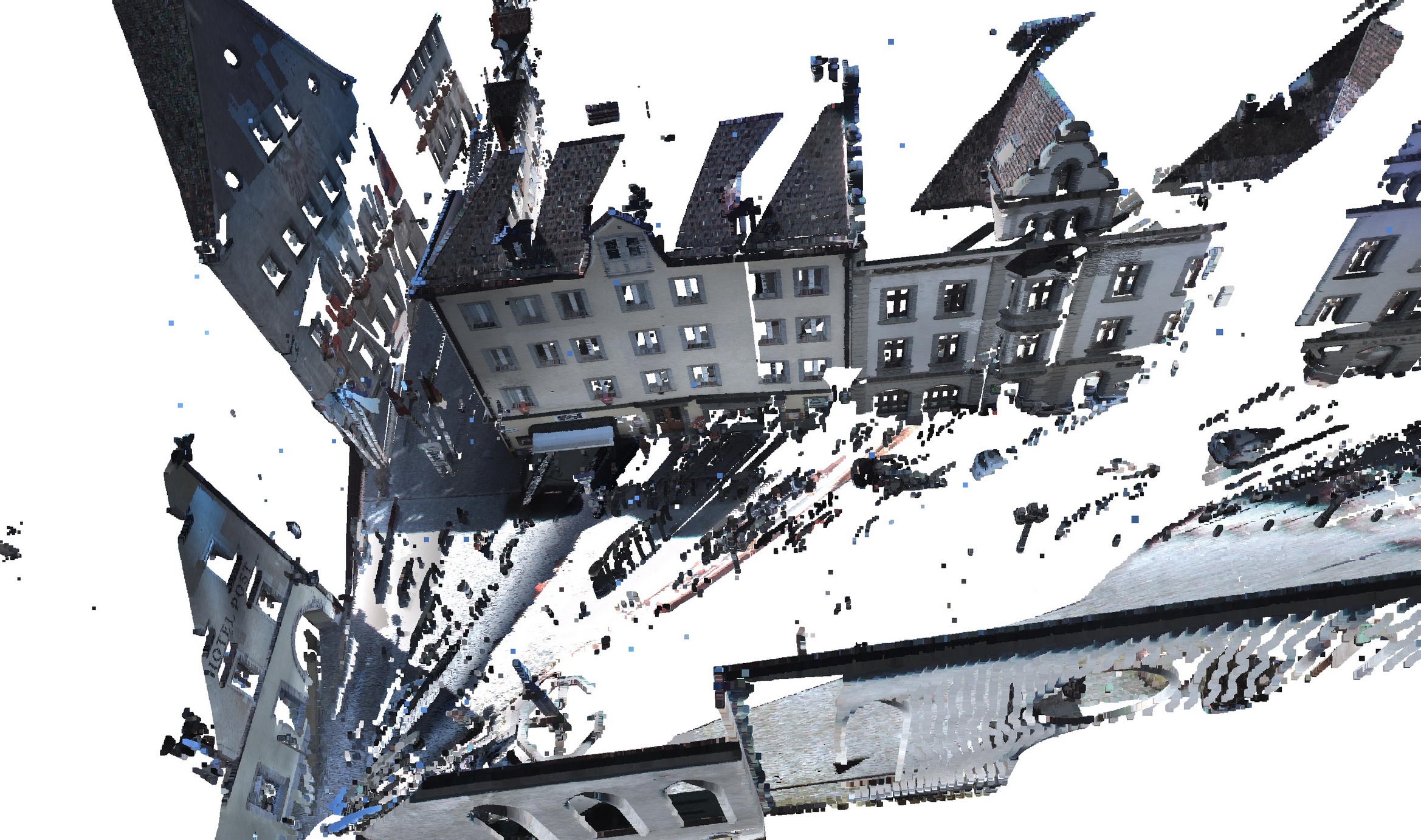}\\
    \includegraphics[width=1\textwidth]{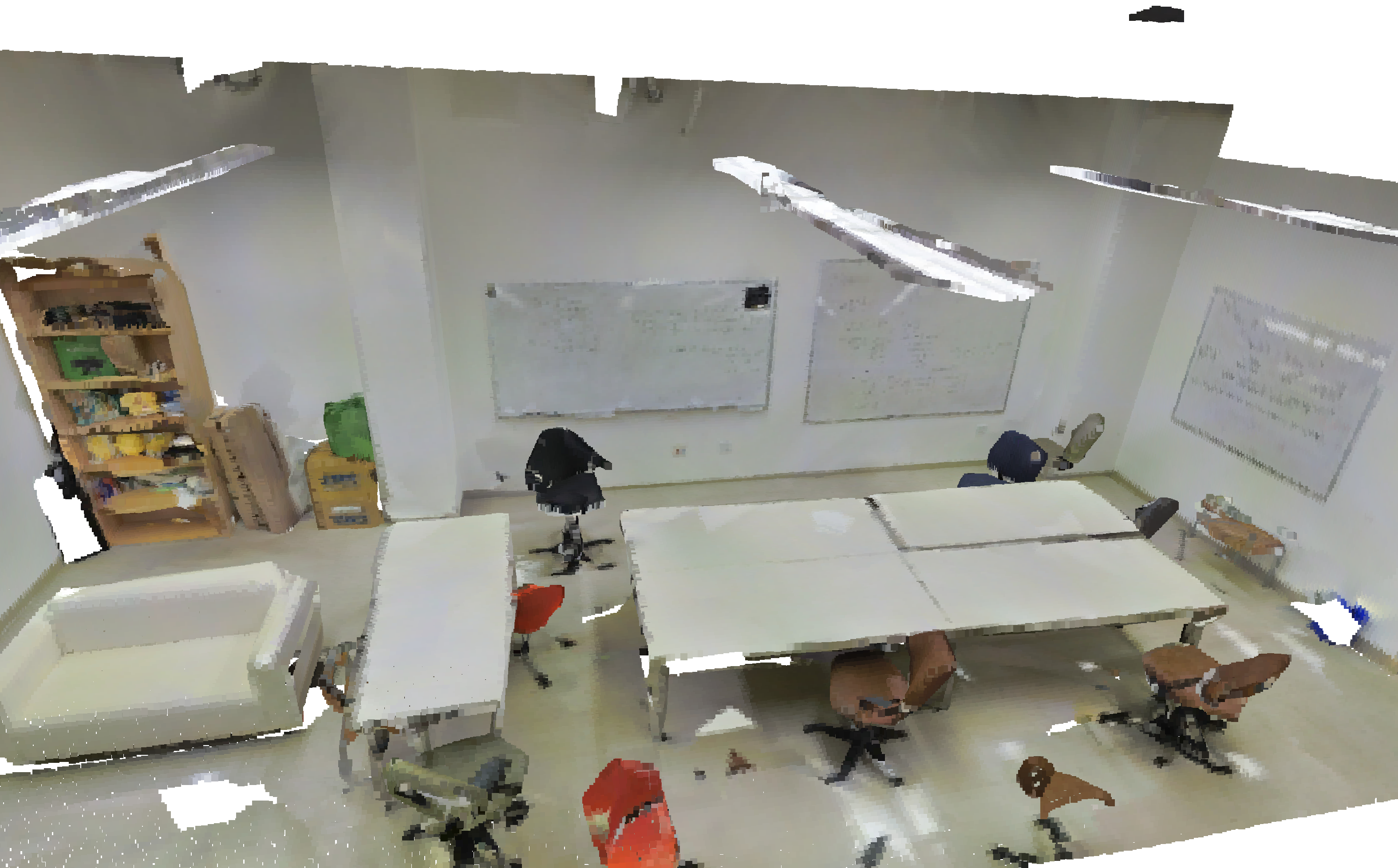}\\ 	  \includegraphics[width=1\textwidth]{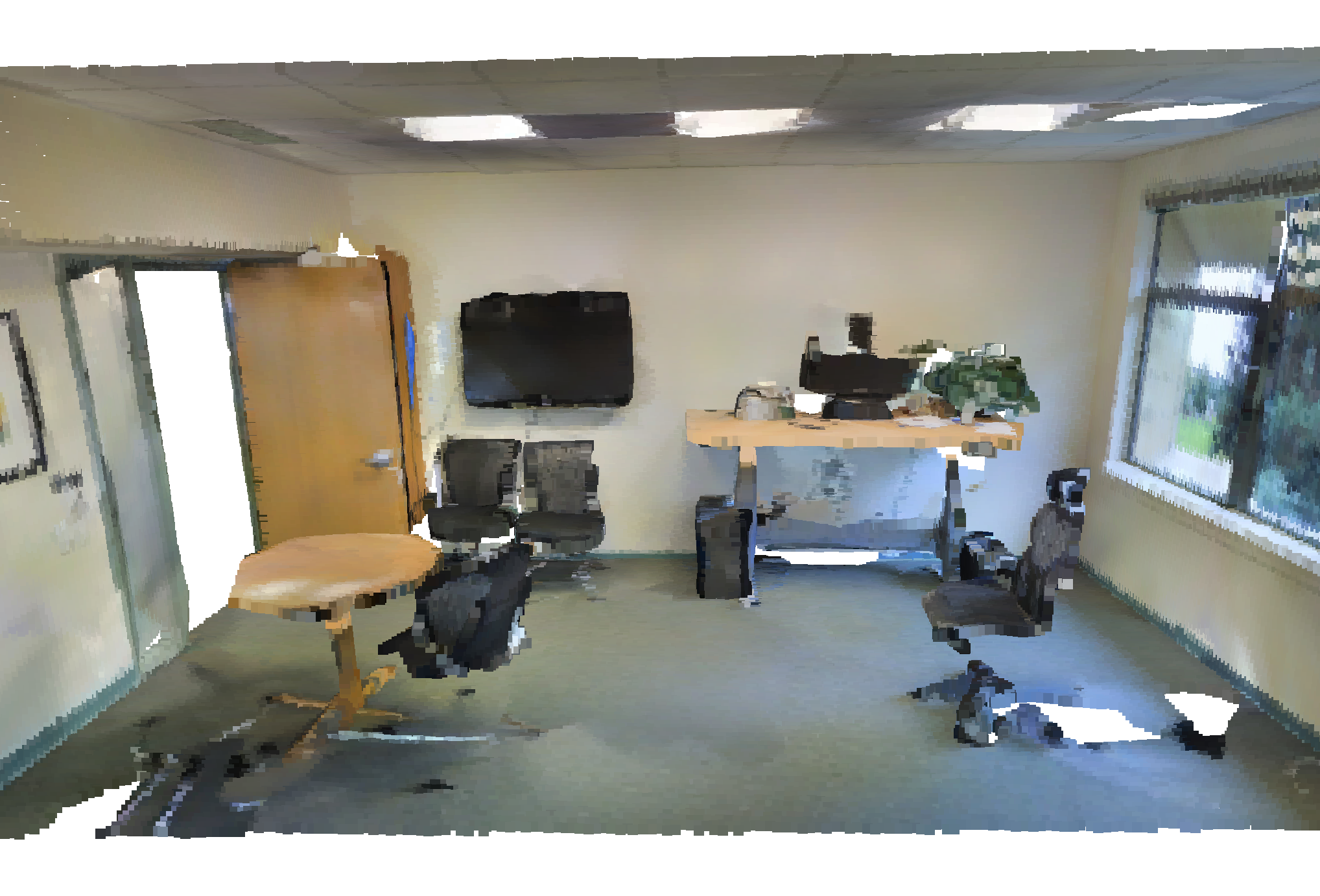}
 \end{tabular}
 \caption{RGB point cloud}
\label{fig:illustration_RGB}
 }\end{subfigure}
 &
  \begin{subfigure}[b]{0.2\textwidth}
  \parbox[b]{1\textwidth}{
 \begin{tabular}[b]{c}
 	\includegraphics[width=1\textwidth]{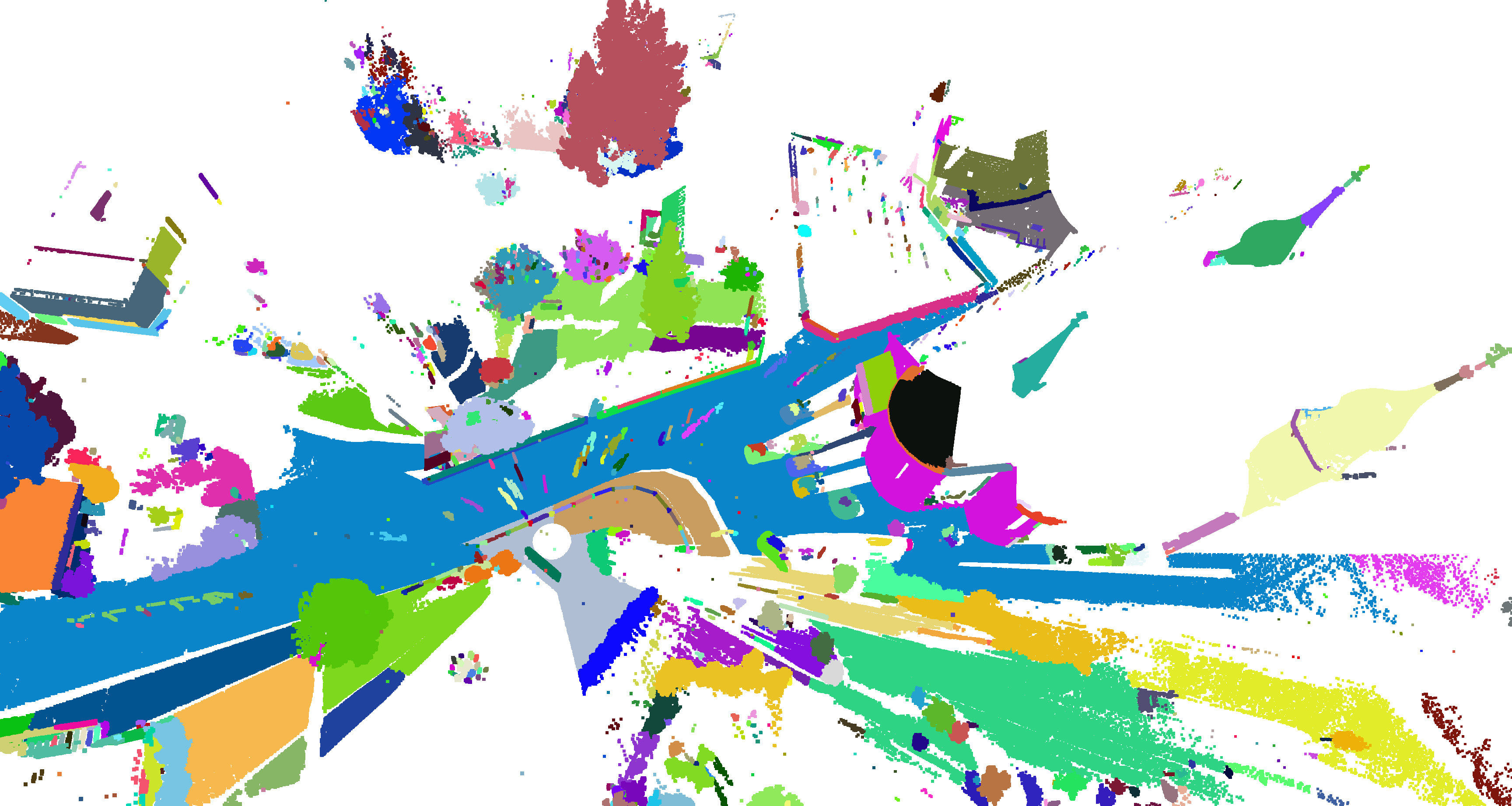}\\
    \includegraphics[width=1\textwidth]{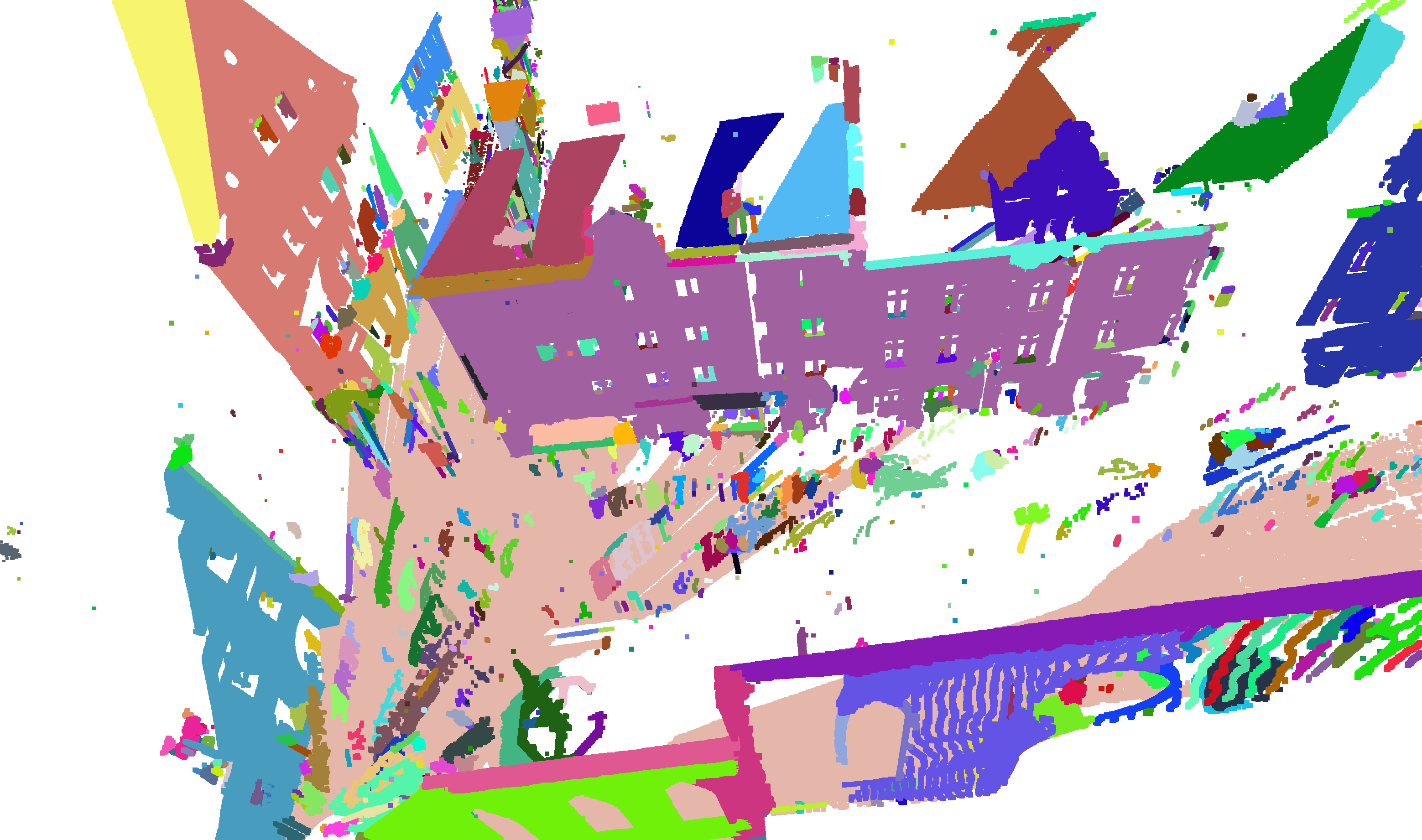}\\
    \includegraphics[width=1\textwidth]{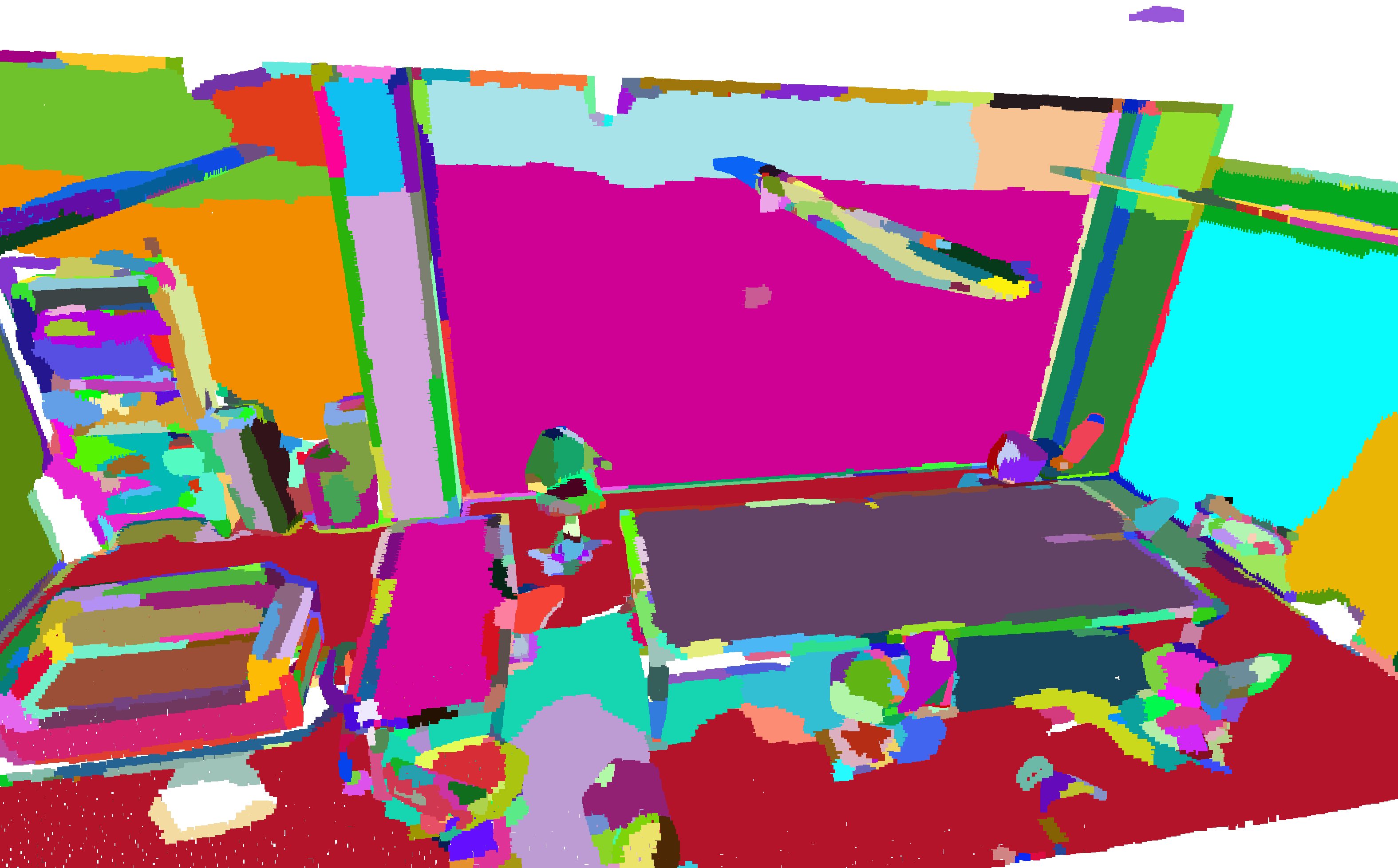}\\ 	  \includegraphics[width=1\textwidth]{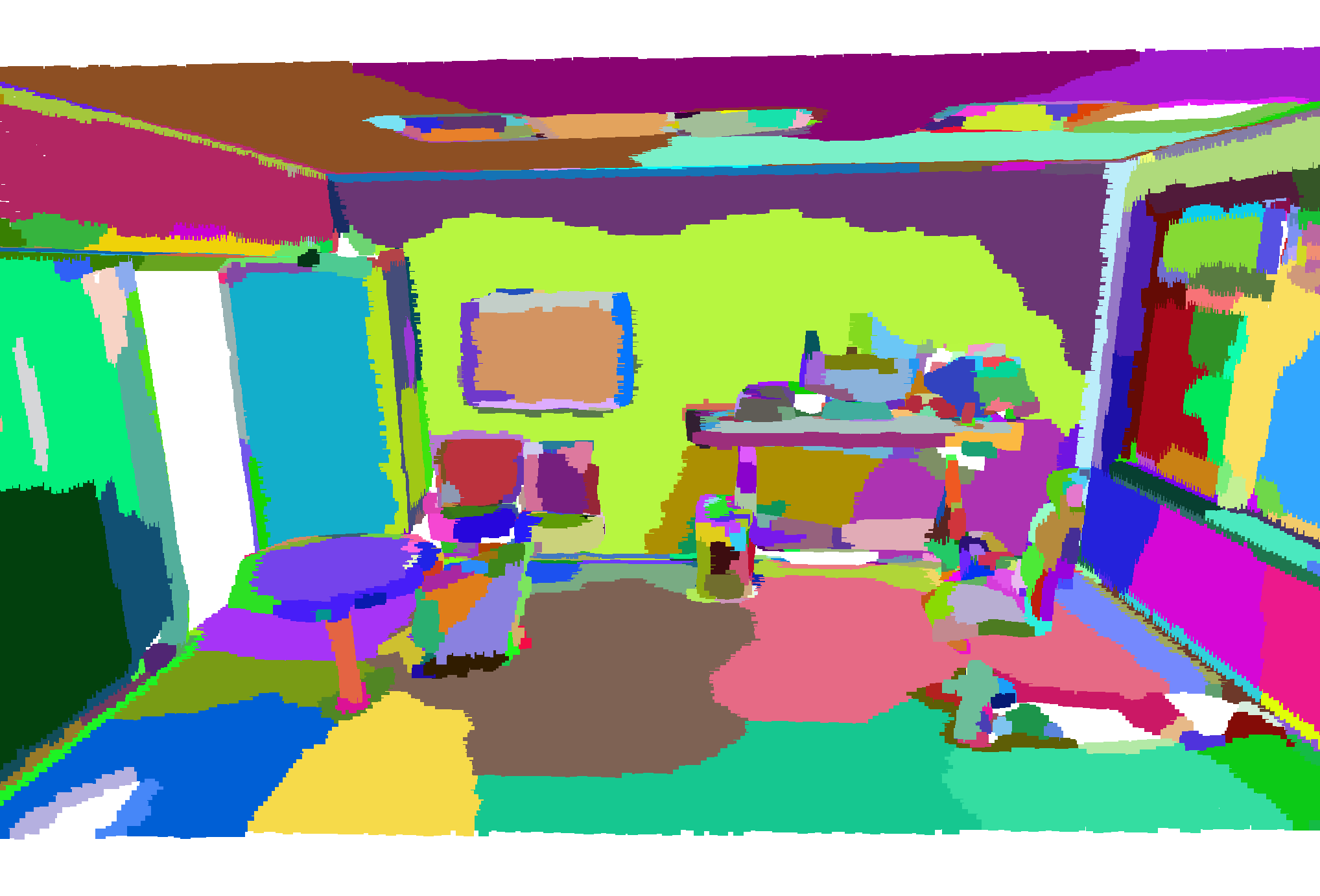}
 \end{tabular}
 \caption{Geometric partitioning}
\label{fig:illustration_seg}
 }\end{subfigure}
 &
  \begin{subfigure}[b]{0.2\textwidth}
  \parbox[b]{1\textwidth}{
 \begin{tabular}[b]{c}
 	\includegraphics[width=1\textwidth]{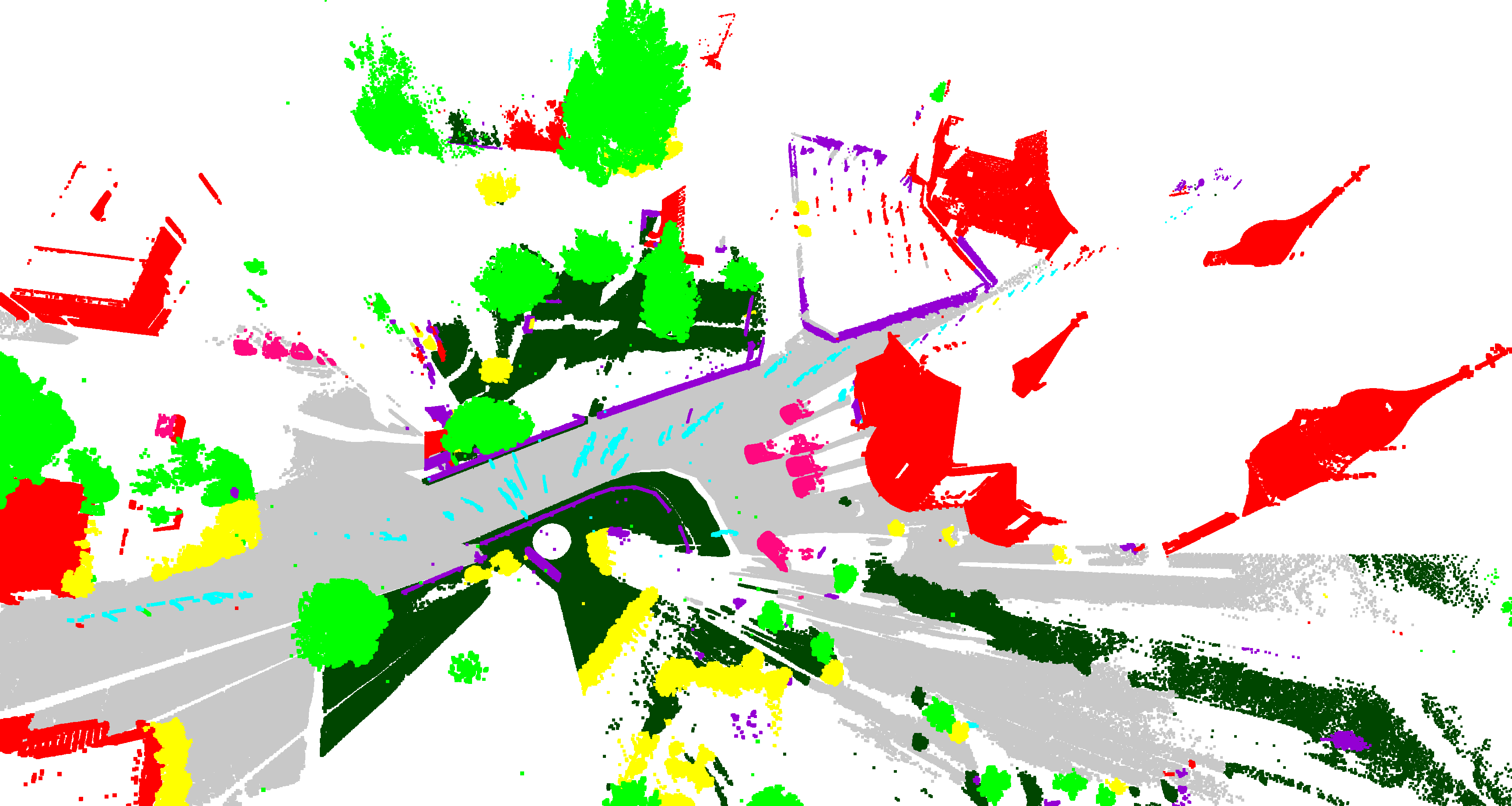}\\
    \includegraphics[width=1\textwidth]{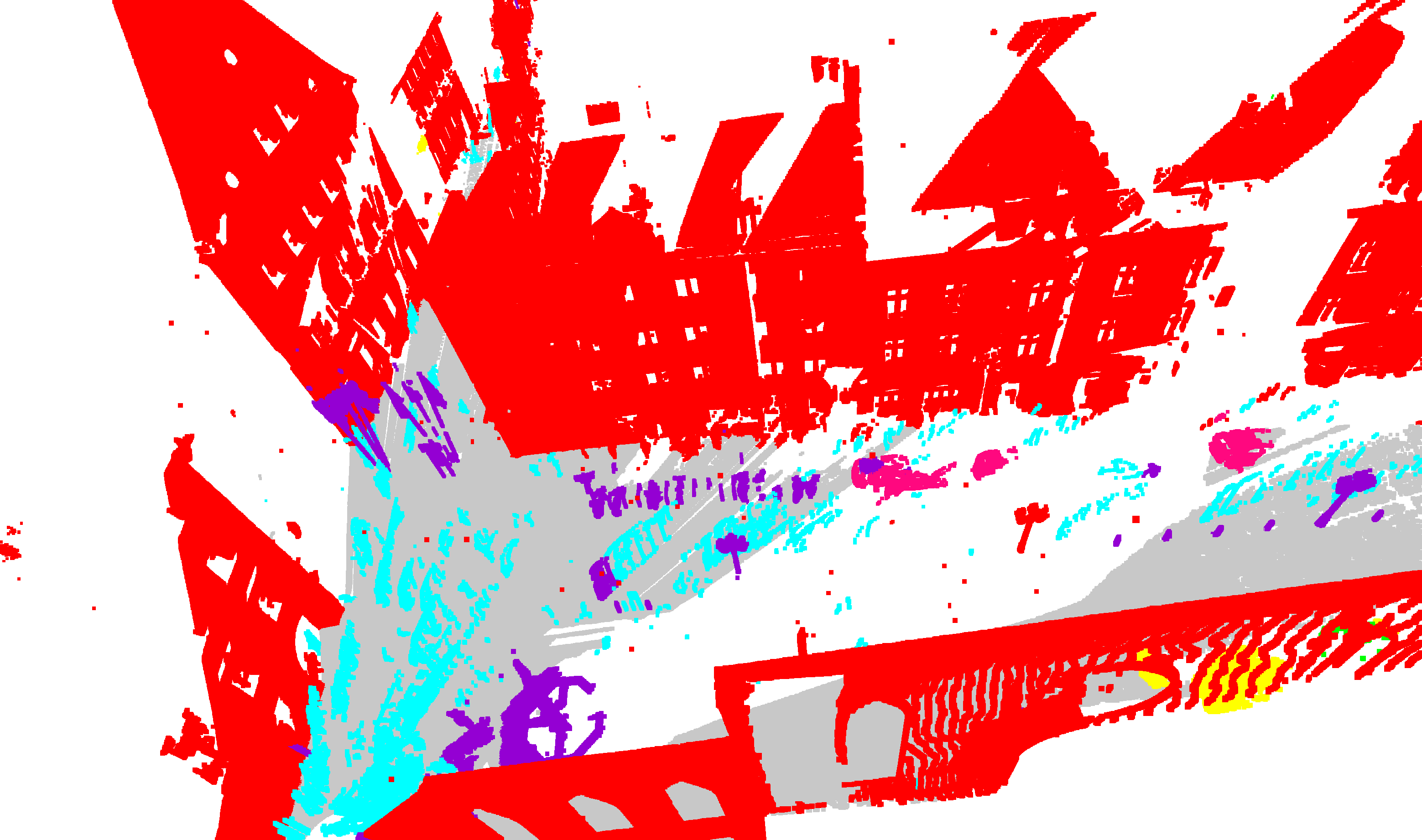}\\
    \includegraphics[width=1\textwidth]{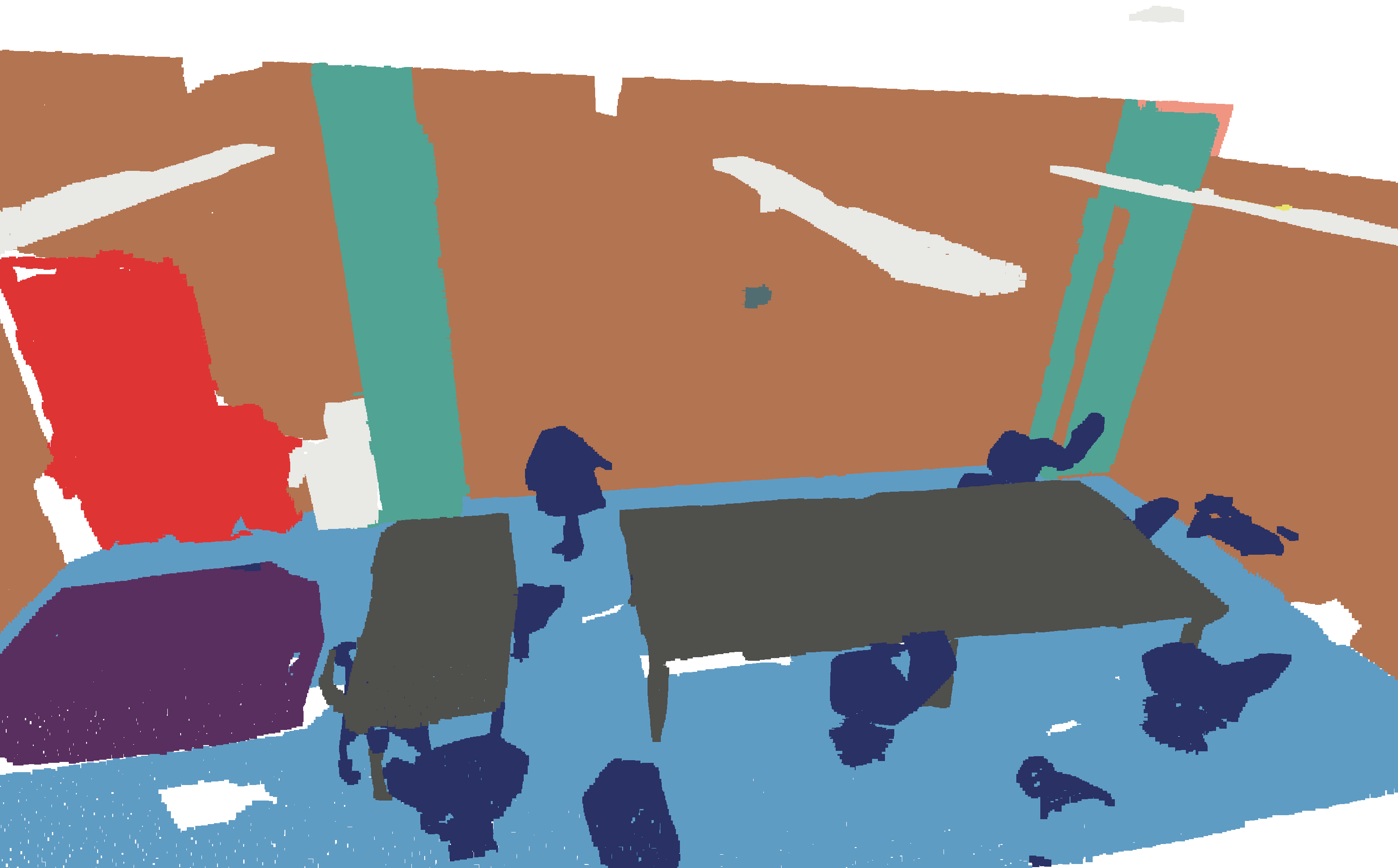}\\ 	  \includegraphics[width=1\textwidth]{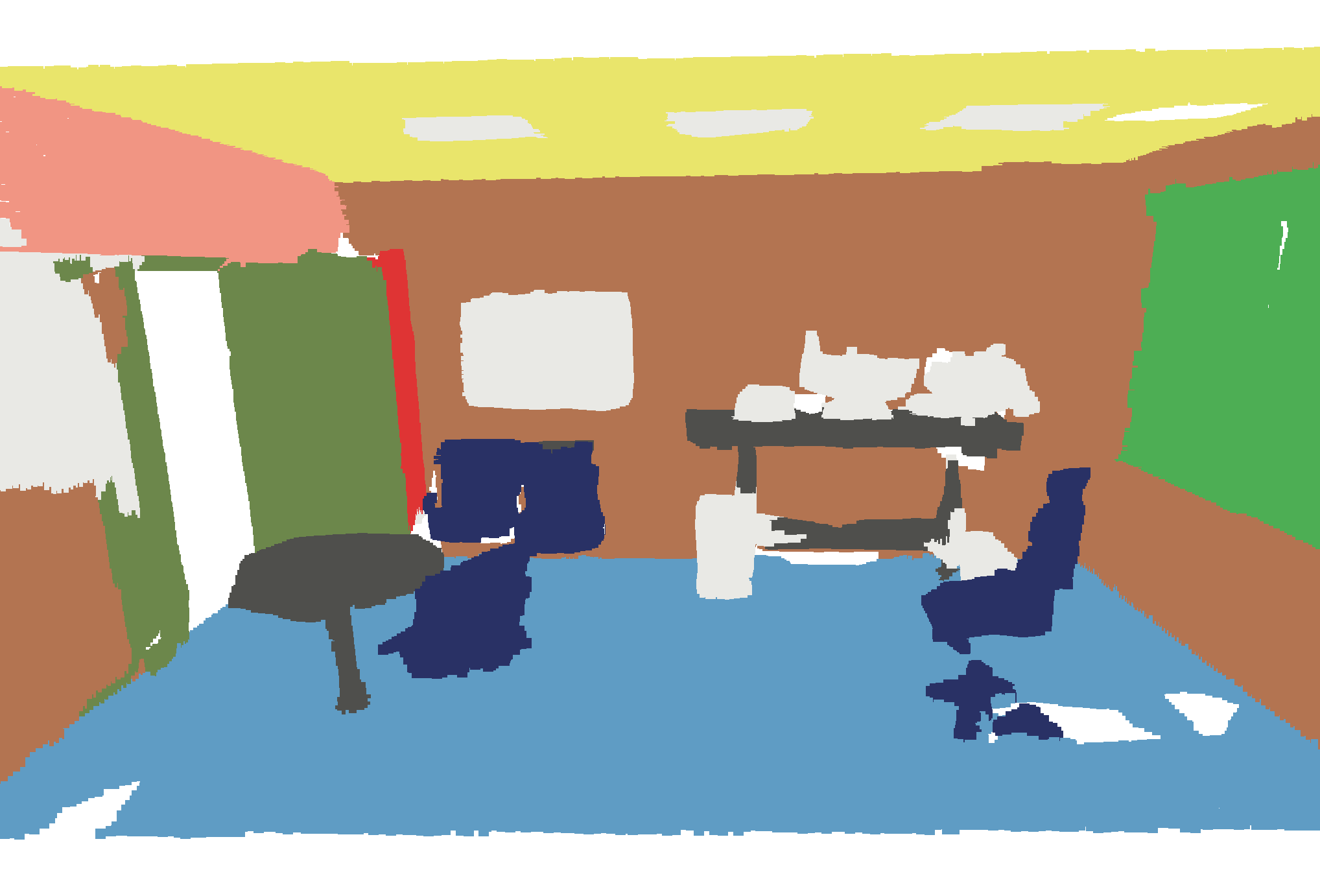}
 \end{tabular}
 \caption{Prediction}
\label{fig:illustration_pred}
 }\end{subfigure}
 &
  \begin{subfigure}[b]{0.2\textwidth}
  \parbox[b]{1\textwidth}{
 \begin{tabular}[b]{c}
 	\includegraphics[width=1\textwidth]{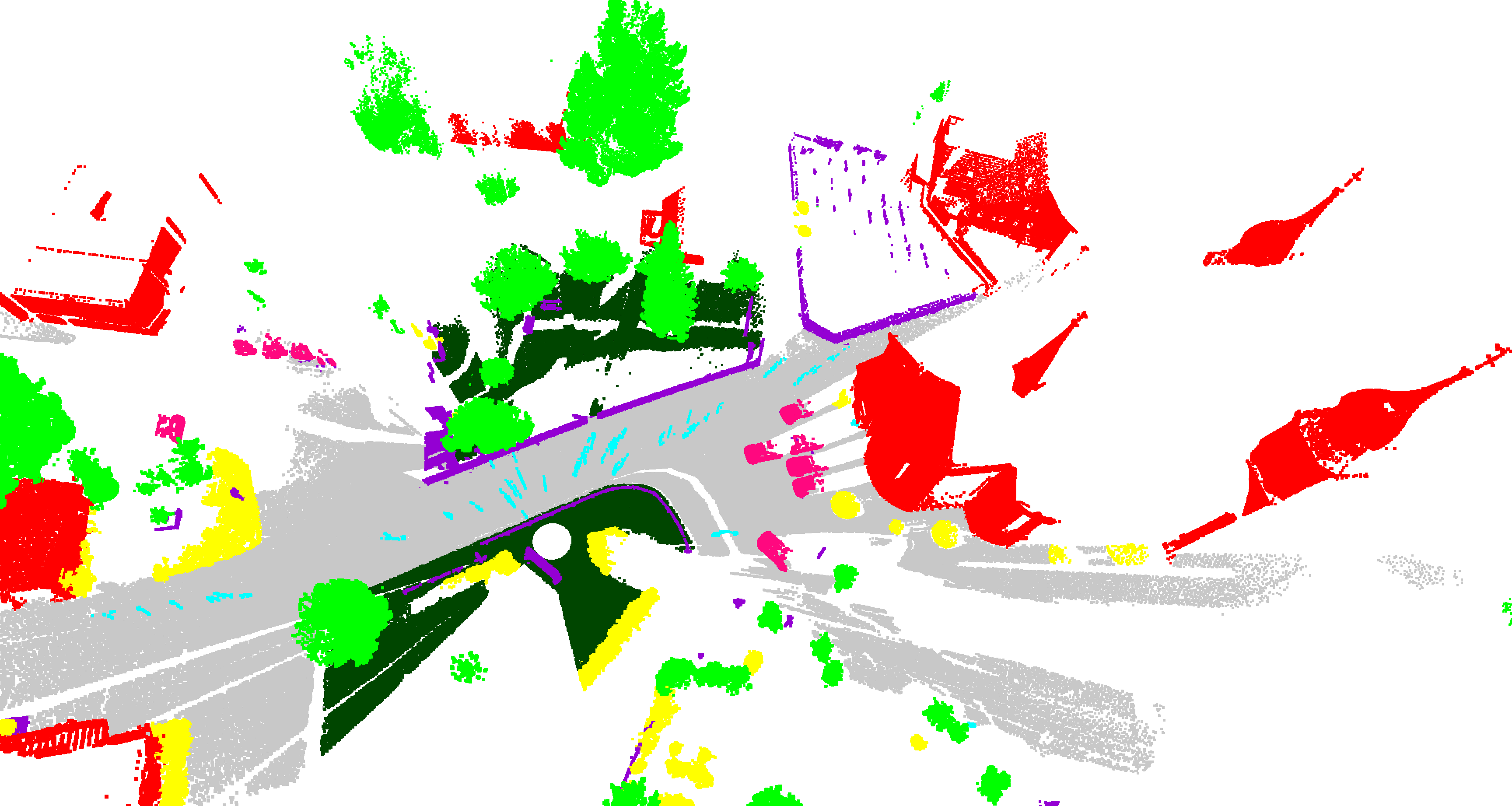}\\
    \includegraphics[width=1\textwidth]{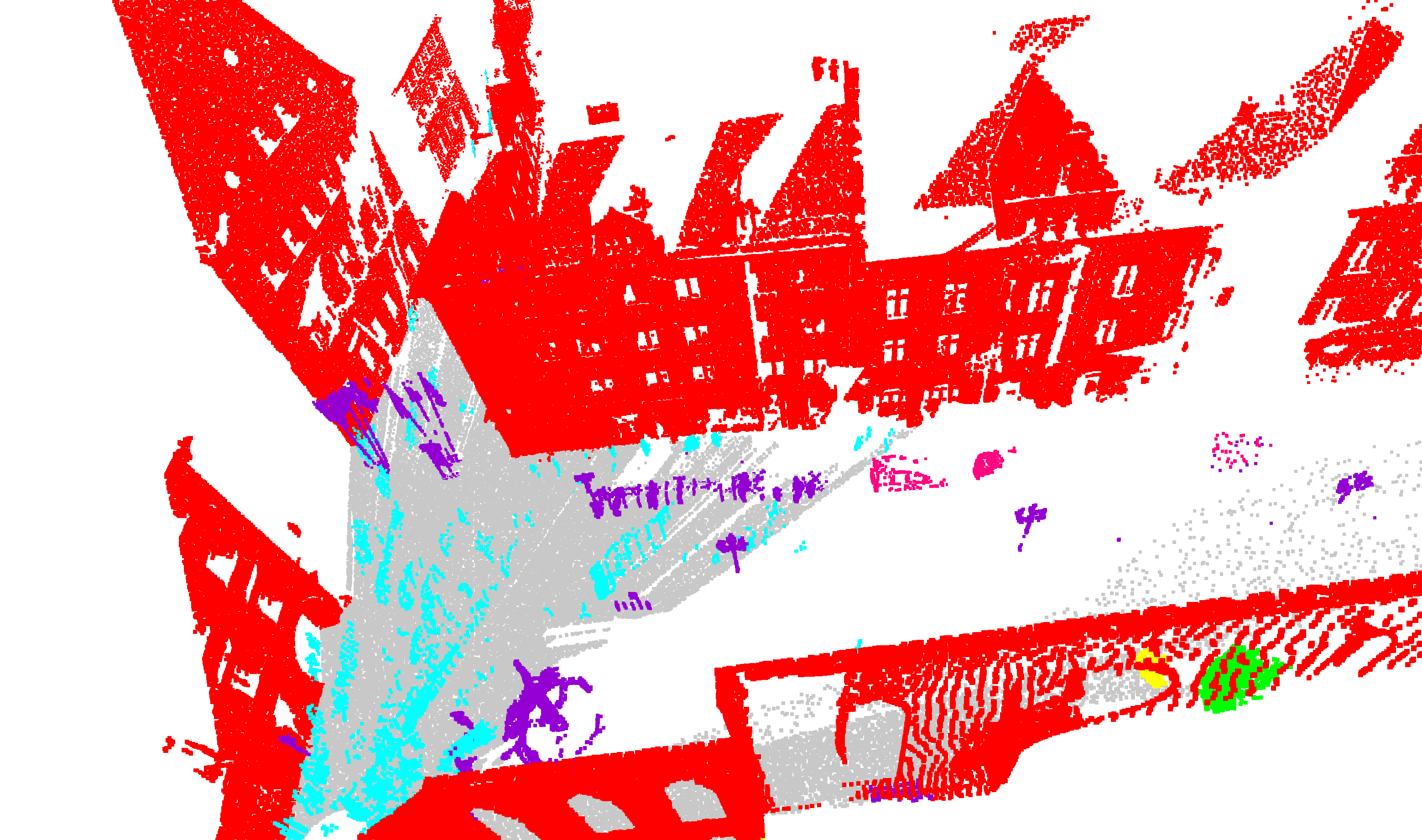}\\
    \includegraphics[width=1\textwidth]{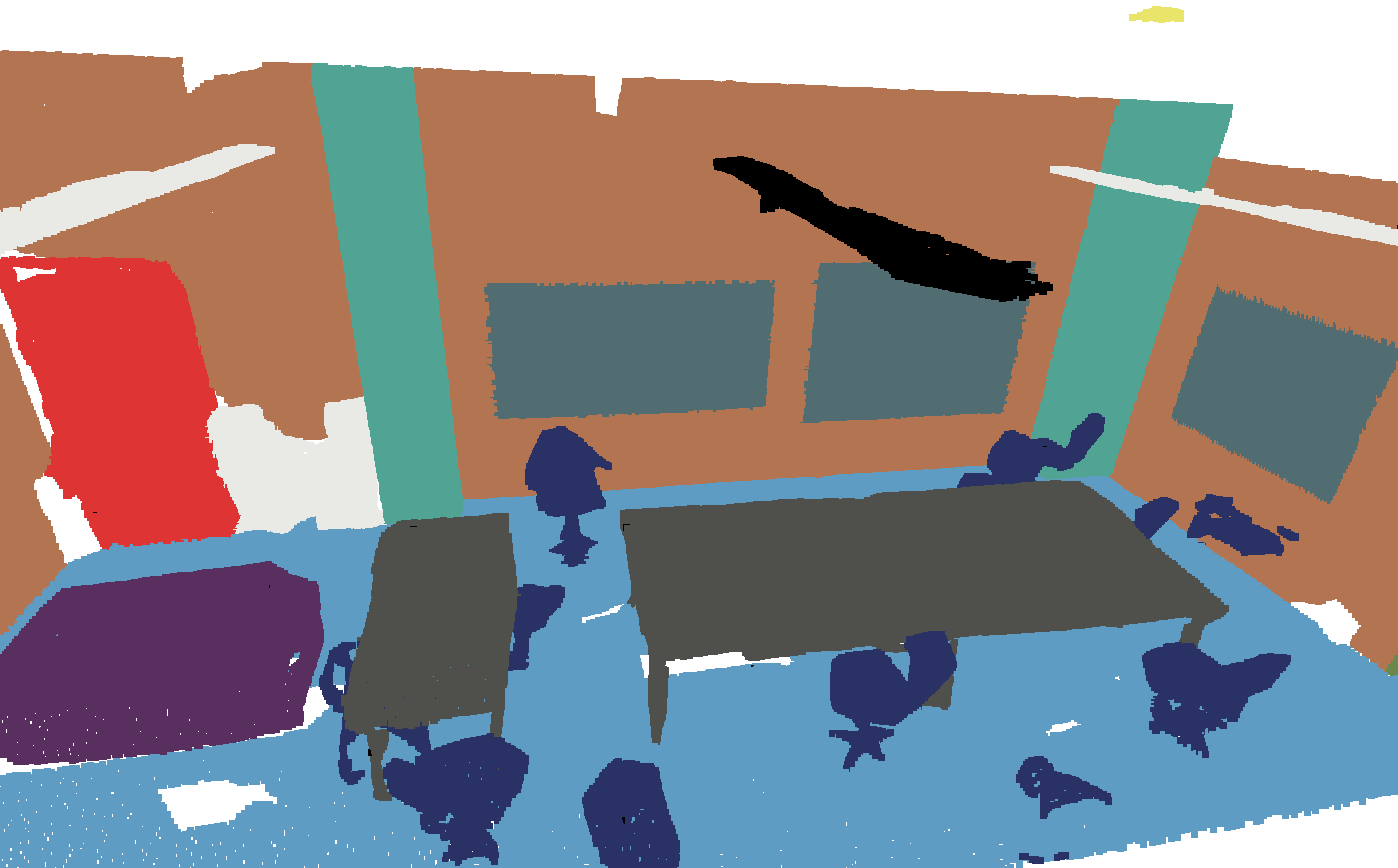}\\ 	  \includegraphics[width=1\textwidth]{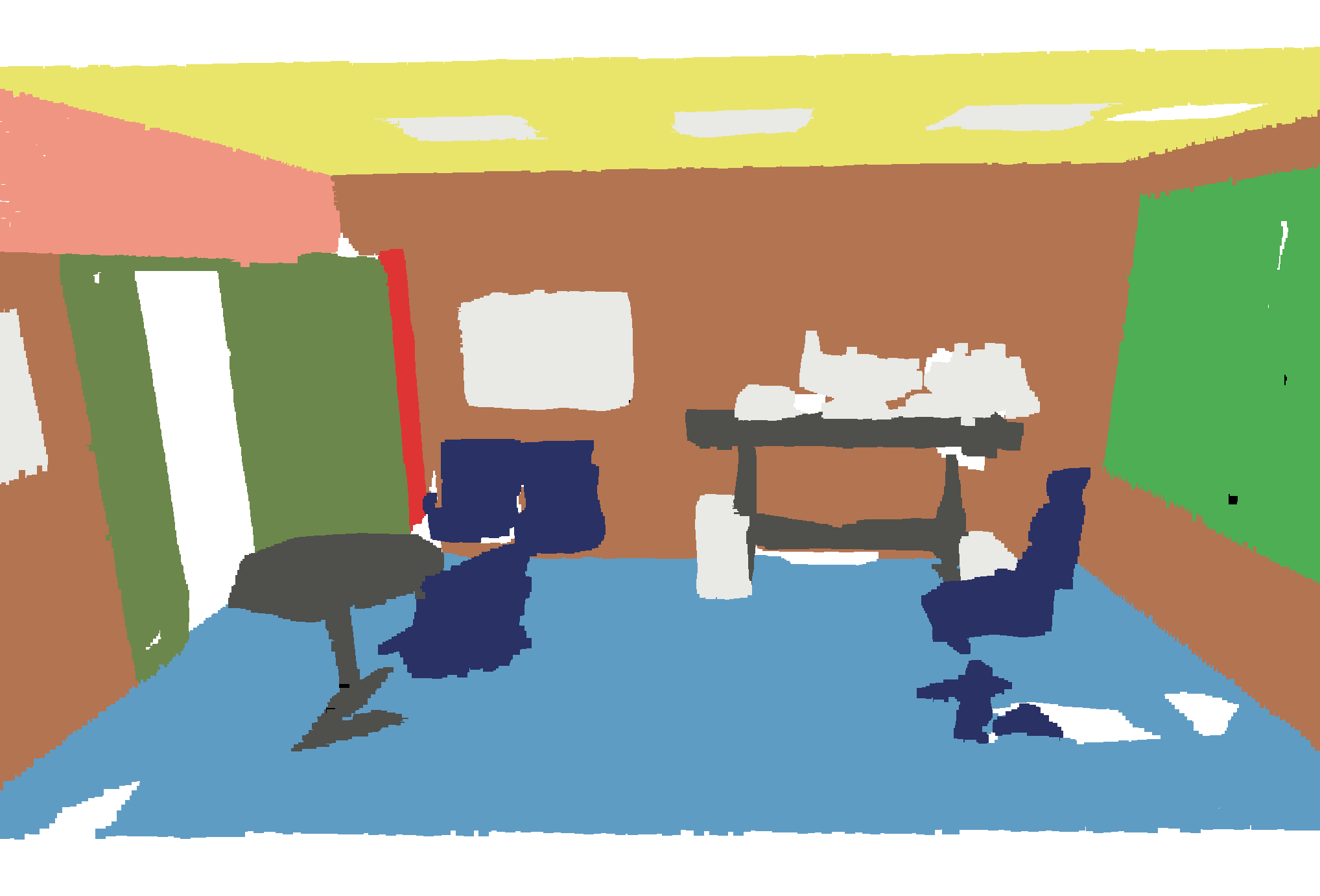}
 \end{tabular}
 \caption{Ground truth}
\label{fig:illustration_GT}
 }\end{subfigure} &
 \begin{subfigure}[t]{0.2\textwidth}
 \parbox[b]{1\textwidth}{
 	%\frame{
 \begin{tabular}[b]{rl}
 \multicolumn{2}{c}{\textbf{Semantic3D}}\\
 \definecolor{tempcolor1}{RGB}{200,200,200}
 \tikz\node at (0,0) [rectangle, minimum width = 3mm, draw = none, fill = tempcolor1] (n){};
 & road\\
 \definecolor{tempcolor2}{RGB}{0,70,0}
 \tikz\node at (0,0) [rectangle, minimum width = 3mm, draw = none, fill = tempcolor2] (n){};
 & grass\\
 \definecolor{tempcolor3}{RGB}{0,255,0}
 \tikz\node at (0,0) [rectangle, minimum width = 3mm, draw = none, fill = tempcolor3] (n){};
 & tree\\
 \definecolor{tempcolor4}{RGB}{255,255,0}
 \tikz\node at (0,0) [rectangle, minimum width = 3mm, draw = none, fill = tempcolor4] (n){};
 & bush\\
 \definecolor{tempcolor5}{RGB}{255,0,0}
 \tikz\node at (0,0) [rectangle, minimum width = 3mm, draw = none, fill = tempcolor5] (n){};
 & buildings\\
 \definecolor{tempcolor6}{RGB}{148,0,211}
 \tikz\node at (0,0) [rectangle, minimum width = 3mm, draw = none, fill = tempcolor6] (n){};
 & hardscape\\
 \definecolor{tempcolor7}{RGB}{0,255,255}
 \tikz\node at (0,0) [rectangle, minimum width = 3mm, draw = none, fill = tempcolor7] (n){};
 & artefacts\\
 \definecolor{tempcolor8}{RGB}{255,8,127}
 \tikz\node at (0,0) [rectangle, minimum width = 3mm, draw = none, fill = tempcolor8] (n){};
 & cars\\
 \multicolumn{2}{c}{\textbf{S3DIS}} \\
 \definecolor{tempcolor9}{RGB}{233,229,107}
 \tikz\node at (0,0) [rectangle, minimum width = 3mm, draw = none, fill = tempcolor9] (n){};
 & ceiling\\
 \definecolor{tempcolor10}{RGB}{95,156,196}
 \tikz\node at (0,0) [rectangle, minimum width = 3mm, draw = none, fill = tempcolor10] (n){};
 & floor\\
 \definecolor{tempcolor11}{RGB}{179,116,81}
 \tikz\node at (0,0) [rectangle, minimum width = 3mm, draw = none, fill = tempcolor11] (n){};
 & wall\\
 \definecolor{tempcolor12}{RGB}{81,163,148}
 \tikz\node at (0,0) [rectangle, minimum width = 3mm, draw = none, fill = tempcolor12] (n){};
 & column\\
 \definecolor{tempcolor13}{RGB}{241,149,131}
 \tikz\node at (0,0) [rectangle, minimum width = 3mm, draw = none, fill = tempcolor13] (n){};
 & beam\\
 \definecolor{tempcolor14}{RGB}{77,174,84}
 \tikz\node at (0,0) [rectangle, minimum width = 3mm, draw = none, fill = tempcolor14] (n){};
 & window\\
 \definecolor{tempcolor15}{RGB}{108,135,75}
 \tikz\node at (0,0) [rectangle, minimum width = 3mm, draw = none, fill = tempcolor15] (n){};
 & door\\
 \definecolor{tempcolor16}{RGB}{79,79,76}
 \tikz\node at (0,0) [rectangle, minimum width = 3mm, draw = none, fill = tempcolor16] (n){};
 & table\\
 \definecolor{tempcolor17}{RGB}{41,49,101}
 \tikz\node at (0,0) [rectangle, minimum width = 3mm, draw = none, fill = tempcolor17] (n){};
 & chair\\
 \definecolor{tempcolor18}{RGB}{223,52,52}
 \tikz\node at (0,0) [rectangle, minimum width = 3mm, draw = none, fill = tempcolor18] (n){};
 & bookcase\\
 \definecolor{tempcolor19}{RGB}{100,20,100}
 \tikz\node at (0,0) [rectangle, minimum width = 3mm, draw = none, fill = tempcolor19] (n){};
& sofa\\
 \definecolor{tempcolor20}{RGB}{81,109,114}
 \tikz\node at (0,0) [rectangle, minimum width = 3mm, draw = none, fill = tempcolor20] (n){};
 & board\\
 \definecolor{tempcolor21}{RGB}{220,220,220}
  \tikz\node at (0,0) [rectangle, minimum width = 3mm, draw = none, fill = tempcolor21] (n){};
 & clutter\\ 
 \definecolor{tempcolor22}{RGB}{0,0,0}
\tikz\node at (0,0) [rectangle, minimum width = 3mm, draw = none, fill = tempcolor22] (n){};
 & unlabelled 
 \end{tabular}%}
 }\end{subfigure}
  \end{tabular}

%% file: supplementary.tex
%=================================================
\section{Model Details}
%=================================================
\label{sec:model_details}

\paragraph*{Voxelization.} 
We pre-process input point clouds with voxelization subsampling by computing per-voxel mean positions and observations over a regular 3D grid ($5$ cm bins for Semantic3D and $3$ cm bins for S3DIS dataset).
The resulting semantic segmentation is interpolated back to the original point cloud in a nearest neighbor fashion.  Voxelization helps decreasing the computation time and memory requirement, and improves the accuracy of the semantic segmentation by acting as a form of geometric and radiometric denoising as well (\ARXIV{\tabref{tab:computation_time}}{Table 4} in the main paper). The quality of further steps is practically not affected, as superpoints are usually strongly subsampled for embedding during learning and inference anyway (\ARXIV{\secref{sec:embedding}}{Subsection 3.3} in the main paper).

\paragraph*{Geometric Partition.} We set regularization strength $\mu=0.8$ for Semantic3D and $\mu=0.03$ for S3DIS, which strikes a balance between semantic homogeneity of superpoints and the potential for their successful discrimination (S3DIS is composed of smaller semantic parts than Semantic3D). In addition to five geometric features $f$ (linearity, planarity, scattering, verticality, elevation), we use color information $o$ for clustering in S3DIS due to some classes being geometrically indistinguishable, such as boards or doors.

\begin{figure}[ht!]
\begin{center}
\resizebox{0.47\textwidth}{!}{ 
    \begin{tikzpicture}[
 triangle/.style = {fill=white, regular polygon, regular polygon sides=3 }]
      \node [rectangle, draw = white, minimum width=2cm,  minimum height=5mm] at (0.75,1.2)   (transformNet)  {\scriptsize STN};
\node [rectangle, draw = black, rotate=90,  minimum width=2cm,  minimum height=5mm] at (0,0)   (input)  {\scriptsize $n_p \times d_p$};
   \node [rectangle, draw = black,  minimum width=.6cm,  minimum height=5mm] at (0.3,-1.5)   (scale)  {\scriptsize diameter};
\node [rectangle, draw = black, rotate=90,  minimum width=2cm,  minimum height=5mm] at (1.5,0)   (transformed)  {\scriptsize $n_p \times d_p$};
  
  \node [rectangle, draw = none, minimum width=2cm,  minimum height=5mm] at (2.8,1.2)   (MLP1)  {\scriptsize MLP(64,64,128,128,256)};
\draw[-, ultra thick, dotted] (0.75, 0.7) -- (0.75, -0.7);
\draw[-, ultra thick, dotted] (2.75, 0.7) -- (2.75, -0.7);
  \node [rectangle, draw = white, minimum width=.5cm,  minimum height=2mm, fill = white] at (0.75,0)   (MLP11)  {\scriptsize shared};
  \node [rectangle, draw = white, minimum width=.5cm,  minimum height=2mm, fill = white] at (2.75,0)   (MLP11)  {\scriptsize shared};
  \node [rectangle, draw = black,  minimum width=1cm,  minimum height=5mm] at (6,0)   (global)  {\scriptsize $257 \times 1$};

  \node [draw = white, rotate=90] at (4,0)   (maxpool)  {\scriptsize maxpool};

  \node [rectangle, draw = white , minimum width=1cm,  minimum height=2mm] at (7,0.5)   (MLP2)  {\scriptsize MLP(256,64,32)};

  \node [rectangle, draw = black,  minimum width=1cm,  minimum height=5mm] at (8,0)   (output)  {\scriptsize $d_z \times 1$};

  \draw[->, ultra thick] (0.25, 0.8) -- (1.25, 0.8);
  \draw[->, ultra thick] (0.25, -0.8) -- (1.25, -0.8);

  \draw[-, ultra thick] (scale) to (3.8, -1.5);
  \draw[->, ultra thick] (3.8, -1.5) to[out = 0, in = 270] (global);

  \draw[->, ultra thick] (1.75, 0.8 ) -- (3.8, 0.8);
  \draw[->, ultra thick] (1.75, -0.8) -- (3.8, -0.8);

  %\draw[thick] (3.2, 1) -- (4.2, 0) -- (3.2, -1) -- (3.2, 1);
  \draw[thick] (3.8, 1) -- (5, 0) -- (3.8, -1) -- (3.8, 1);

 \draw[->, ultra thick] (5, 0) -- (5.5, 0);

 %\draw[->, ultra thick] (5.5, 0) -- (6, 0);
 \draw[->, ultra thick] (6.5, 0) -- (7.5, 0);

    \end{tikzpicture}
}
\end{center}
\caption{The PointNet embedding $n_p$ $d_p$-dimensional samples of a superpoint to a $d_z$-dimensional vector.}
\label{ref:fig_pointnet}
\end{figure}
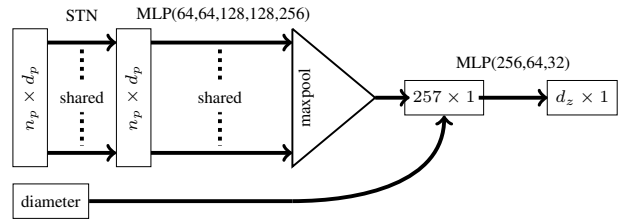

\paragraph*{PointNet.} We use a simplified shallow and narrow PointNet architecture with just a single Spatial Transformer Network (STN), see \figref{ref:fig_pointnet}. We set $n_p=128$ and $n_\mathrm{minp}=40$. Input points are processed by a sequence of MLPs (widths 64, 64, 128, 128, 256) and max pooled to a single vector of 256 features. The scalar metric diameter is appended and the result further processed by a sequence of MLPs (widths 256, 64, $d_z$=32). A residual matrix $\Phi \in \mathbb{R}^{2\times 2}$ is regressed by STN and $(I+\Phi)$ is used to transform XY coordinates of input points as the first step. The architecture of STN is a "small PointNet" with 3 MLPs (widths 64, 64, 128) before max pooling and 3 MLPs after (widths 128, 64, 4). Batch Normalization~\cite{IoffeS15_bn} and ReLUs are used everywhere. Input points have $d_p$=11 dimensional features for Semantic3D (position $p_i$, color $o_i$, geometric features $f_i$), with 3 additional ones for S3DIS (room-normalized spatial coordinates, as in past work \cite{qi2016pointnet}).

\paragraph*{Segmentation Network.} We use embedding dimensionality $d_z=32$ and $T=10$ iterations. ECC-VV is used for Semantic3D (there are only $15$ point clouds even though the amount of points is large), while ECC-MV is used for S3DIS (large number of point clouds).
Filter-generating network $\Theta$ is a MLP with 4 layers (widths 32, 128, 64, and 32 or $32^2$ for ECC-VV or ECC-MV) with ReLUs. Batch Normalization is used only after the third parametric layer. No bias is used in the last layer. Superedges have $d_f=13$ dimensional features, normalized by mean subtraction and scaling to unit variance based on the whole training set.

\paragraph*{Training.} We train using Adam~\cite{KingmaB14_adam} with initial learning rate 0.01 and batch size 2, \ie effectively up to 1024 superpoints per batch. For Semantic3D, we train for 500 epochs with stepwise learning rate decay of 0.7 at epochs 350, 400, and 450. For S3DIS, we train for 250 epochs with steps at 200 and 230. We clip gradients within $[-1,1]$.

%=================================================
\section{CRF-ECC}
%=================================================
\label{sec:crfecc}

In this section, we describe our adaptation of CRF-RNN mean field inference by Zheng \etal~\cite{Zheng15crf} for post-processing PointNet embeddings in SPG, denoted as unary potentials $U_i$ here. 

The original work proposed a dense CRF with pairwise potentials $\Psi$ defined to be a mixture of $m$ Gaussian kernels as $\Psi_{ij} = \mu \sum_m w_m K_m(F_{ij})$, where $\mu$ is label compatibility matrix, $w$ are parameters, and $K$ are fixed Gaussian kernels applied on edge features. 

We replace this definition of the pairwise term with a Filter generating network $\Theta$~\cite{simonovsky2017dynamic} parameterized with weights $W_e$, which generalizes the message passing and compatibility transform steps of Zheng \etal. Furthermore, we use superedge connectivity $\cE$ instead of assuming a complete graph. The pseudo-code is listed in Algorithm~\ref{alg:crfecc}. Its output are marginal probability distributions $Q$. In practice we run the inference for $T=10$ iterations.

\begin{algorithm}
\caption{CRF-ECC}\label{alg:crfecc}
\begin{algorithmic}
\State $Q_i \gets \mathrm{softmax}(U_i)$
\While{not converged}
  \State $\hat{Q}_i \gets \sum_{j \mid (j,i) \in \cE} \Theta(F_{ji,\cdot}; W_e) Q_j$
  \State $\breve{Q}_i \gets U_i - \hat{Q}_i$
  \State $Q_i \gets \mathrm{softmax}(\breve{Q}_i)$
\EndWhile
\end{algorithmic}
\end{algorithm}
%=================================================
\section{Extended Ablation Studies}
%=====
\label{sec:ext_ablation}

In this section, we present additional set of experiments to validate our design choices and present their results in \tabref{tab:results}.

\textbf{a) Spatial Transformer Network.} While STN makes superpoint embedding orientation invariant, the relationship with surrounding objects are still captured by superedges, which are orientation variant. In practice, STN helps by $4$ mIoU points.

\textbf{b) Geometric Features.} Geometric features $f_i$ are computed in the geometric partition step and can therefore be used in the following learning step for free. While PointNets could be expected to learn similar features from the data, this is hampered by superpoint subsampling, and therefore their explicit use helps (+4 mIoU).

\textbf{c) Sampling Superpoints.} The main effect of subsampling SPG is regularization by data augmentation. Too small a sample size leads to disregarding contextual information (-4 mIoU) while too large a size leads to overfitting (-2 mIoU). Lower memory requirements at training is an extra benefit. There is no subsampling at test time.

\textbf{d) Long-range Context.} We observe that limiting the range of context information in SPG harms the performance. Specifically, capping distances in $G_{\text{vor}}$ to $1$ m (as used in PointNet~\cite{qi2016pointnet}) or $5$ m (as used in SegCloud\footnote{Furthermore, SegCloud divides the inference into cubes without overlap, possibly causing inconsistencies across boundaries.}~\cite{tchapmi2017segcloud}) worsens the performance of our method (even more on our Semantic 3D validation set).

\textbf{e) Input Gate.} We evaluate the effect of input gating (IG) for GRUs as well as LSTM units. While a LSTM unit achieves higher score than a GRU (-3 mIoU), the proposed IG reverses this situation in favor of GRU (+1 mIoU). Unlike the standard input gate of LSTM, which controls the information flow from the hidden state and input to the cell, our IG controls the input even before it is used to compute all other gates.

\textbf{f) Regularization Strength $\mu$.} We investigate the balance between superpoints' discriminative potential and their homogeneity controlled by parameter $\mu$ . We observe that the system is able to perform reasonably over a range of SPG sizes. 

\textbf{g) Superpoint Sizes.}
We include a breakdown of superpoint sizes for $\mu=0.03$ in relation to hyperparameters $n_\mathrm{minp}=40$ and $n_p=128$, showing that 93\% of points are in embedded superpoints,  and 79\% in superpoints that are subsampled.

\textbf{Superedge Features.} Finally, in \tabref{tab:edge_ablation} we evaluate empirical importance of individual superedge features by removing them from $\mathrm{Best}$. Although no single feature is crucial, the most being offset deviation (+3 mIoU), we remind the reader than without any superedge features the network performs distinctly worse ($\mathrm{NoEdgeFeat}$, -22 mIoU).

\begin{table}\small\begin{center}
\resizebox{1\linewidth}{!}{
\begin{tabular}{|c|cccc|} \hline
\emph{a) Spatial transf.} & no &\bf yes & &\\
mIoU & 58.1 & 62.1 & & \\\hline
\emph{b) Geometric features} & no &\bf yes & &\\
mIoU & 58.4  & 62.1 & & \\\hline
\emph{c) Max superpoints} & 256 & \bf 512 & 1024 &\\
mIoU & 57.9 & 62.1 & 60.4 &\\\hline
\emph{d) Superedge limit} & 1 m & 5 m & \boldmath$\infty$ & \\
mIoU & 61.0 & 61.3 & 62.1 &\\\hline
\emph{e) Input gate} & LSTM & LSTM+IG & GRU & \bf GRU+IG \\
mIoU & 61.0 & 61.0 & 57.5 & 62.1 \\\hline
\emph{f) Regularization $\mu$} & 0.01 & 0.02 & \bf 0.03 & 0.04 \\
\# superpoints & 785 010& 385 091 & 251 266& 186 108 \\
perfect mIoU & 90.6& 88.2  & 86.6 & 85.2\\
mIoU & 59.1 & 59.2 &   62.1&  58.8 \\\hline
\emph{g) Superpoint size} & 1-40 & 40-128 & 128-1000 & $\geq$ 1000 \\
proportion of points& 7\% & 14\% & 27\% & 52\% \\\hline
\end{tabular}}\end{center}
\caption{Ablation study of design decisions on S3DIS (6-fold cross validation). Our choices in bold.}
\label{tab:results}
\vspace*{-0.3cm}
\end{table} 

\begin{table}\begin{center}\small
\begin{tabular}{|c|cc|}\hline
Model & mAcc & mIoU\\\hline
$\mathrm{Best}$ & 73.0 & 62.1\\
no mean offset & 72.5 & 61.8\\
no offset deviation & 71.7 & 59.3\\
no centroid offset & 74.5 & 61.2\\
no len/surf/vol ratios & 71.2 & 60.7\\
no point count ratio & 72.7 & 61.7 \\
\hline
\end{tabular}
\end{center}
\caption{Ablation study of superedge features on S3DIS (6-fold cross validation).}
\label{tab:edge_ablation}
\end{table}
%==================================================
\begin{figure}
\begin{tikzpicture} \begin{axis}[xmin=2, xmode=log, enlargelimits=false, log ticks with fixed point, xtick={1, 40, 128, 1000, 10000}, xlabel=size of superpoints, ylabel = number of points] 
\addplot [const plot,fill=blue,draw=black] 
coordinates {(1,10378) (2,185078) (5,1222298) (15,1552481)(26,2264432) (47,2585369) (86,2574817) (156,2501497) (282,2359433) (510,2614968) (923,3305617)
(1669,3684039) (3018,3708385) (5458,2241618) (9869,1314260)
(17850,369831) (32740,77328)
} \closedcycle;
\addplot[color=red, ultra thick] coordinates { (40,10378) (40,2264432)};
\addplot[color=red, ultra thick] coordinates { (128,10378) (128,2574817)};
\end{axis} \end{tikzpicture} 
\caption{Histogram of points contained in superpoints of different size (in log scale) on the full S3DIS dataset. The embedding threshold $n_\mathrm{minp}$ and subsampling threshold $n_p$ are marked in red.}
\end{figure}
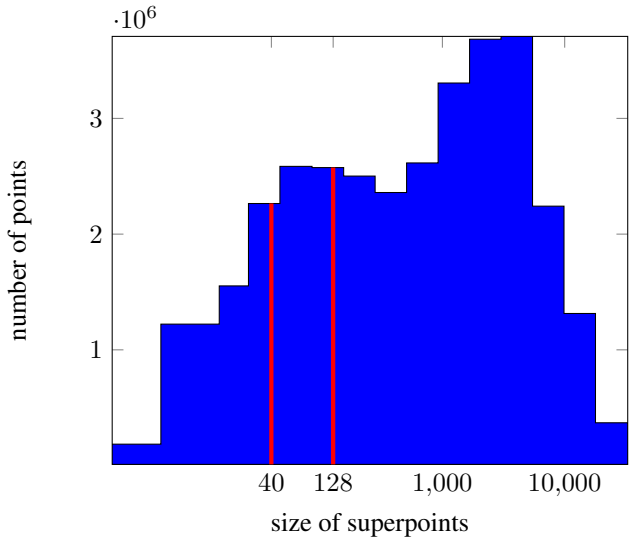
%=================================================
\section{Video Illustration}
%=================================================
\label{sec:video}
We provide a video illustrating our method and qualitative results on S3DIS dataset, which can be viewed at \url{https://youtu.be/Ijr3kGSU_tU}.